\documentclass[runningheads]{llncs}
\usepackage{graphicx}
\usepackage{comment}
\usepackage{amsmath,amssymb} 
\usepackage{xcolor,booktabs}
\usepackage{multirow}
\usepackage{adjustbox}

\usepackage{amsmath,amsfonts,bm}

\def\eqref#1{equation~\ref{#1}}

\def\1{\bm{1}}

\newcommand{\test}{\mathcal{D_{\mathrm{test}}}}

\DeclareMathAlphabet{\mathsfit}{\encodingdefault}{\sfdefault}{m}{sl}
\SetMathAlphabet{\mathsfit}{bold}{\encodingdefault}{\sfdefault}{bx}{n}

\newcommand{\E}{\mathbb{E}}

\newcommand{\KL}{D_{\mathrm{KL}}}

\usepackage{pifont}
\newcommand{\xmark}{\ding{55}}%

\usepackage[width=122mm,left=12mm,paperwidth=146mm,height=193mm,top=12mm,paperheight=217mm]{geometry}

\begin{document}
\pagestyle{headings}
\mainmatter
\def\ECCVSubNumber{5448}  

\title{OneGAN: Simultaneous Unsupervised Learning of Conditional Image Generation, Foreground Segmentation, and Fine-Grained Clustering} 

\titlerunning{OneGAN}
%
\author{Yaniv Benny\inst{1} \and
Lior Wolf\inst{1,2}}
\authorrunning{Y. Benny and L. Wolf}
%
\institute{Tel-Aviv University, Israel \and
Facebook AI Research}
\maketitle

\begin{abstract}
We present a method for simultaneously learning, in an unsupervised manner, (i) a conditional image generator, (ii) foreground extraction and segmentation, (iii) clustering into a two-level class hierarchy, and (iv) object removal and background completion, all done without any use of annotation. The method combines a Generative Adversarial Network and a Variational Auto-Encoder, with multiple encoders, generators and discriminators, and benefits from solving all tasks at once. The input to the training scheme is a varied collection of unlabeled images from the same domain, as well as a set of background images without a foreground object. In addition, the image generator can mix the background from one image, with a foreground that is conditioned either on that of a second image or on the index of a desired cluster. The method obtains state of the art results in comparison to the literature methods, when compared to the current state of the art in each of the tasks.
\end{abstract}

\section{Introduction}
We hypothesize that solving multiple unsupervised tasks together, enables one to improve on the performance of the best methods that solve each individually. The underlying motivation is that in unsupervised learning, the structure of the data is a key source of knowledge and each task exposes a different aspect of it. We advocate for solving the various tasks in phases, where easier tasks are addressed first, and the other tasks are introduced gradually, while constantly updating the solutions of the previous sets of tasks.
The method consists of multiple networks that are trained end-to-end and side-by-side to solve multiple tasks. The method starts from learning background image synthesis and image generation of objects from a particular domain. It then advances to more complex tasks, such as clustering, semantic segmentation and object removal. Finally, we show the model's ability to perform image-to-image translation. The entire learning process is unsupervised, meaning that no annotated information is used. In particular, the method does not employ class labels, segmentation masks, bounding boxes, etc. However, it does require a separate set of clean background images, which are easy to obtain in many cases.

\noindent{\bf Contributions\quad} Beyond the conceptual novelty of a method that treats single-handedly multiple unsupervised tasks, which were previously solved by individual methods, the method displays a host of technical novelties, including: (i) a novel architecture that supports multiple paths addressing multiple tasks, (ii) employing bypass paths that allow a smooth transition between autoencoding and generation based on a random seed, (iii) backpropagation through three paths in each iteration, (iv) mixup module, which applies interpolation between latent representations of the generation and reconstruction paths, and more. Due to each of these novelties, backed by the ablation studies, we obtain state of the art results compared to the literature methods in each of the individual tasks.

\section{Related work}
Since our work touches on many tasks, we focus the literature review on general concepts and on the most relevant work. \textbf{Generative models} are typically based on Generative Adversarial Networks~\cite{goodfellow2014generative} or Variational Auto-Encoders~\cite{vae}. In addition, these two can be combined~\cite{larsen2015autoencoding}. 
\textbf{Conditional image generation}
conditions the output on an initial variable, most commonly, the target class.
CGAN~\cite{mirza2014conditional} and InfoGAN~\cite{chen2016infogan} proposed different methods to apply the condition on the discriminator. Our work is more similar to InfoGAN, since we do not use labeled data and the label is not linked to any real image and no conditional discriminator can be applied. The condition is maintained by a classifier that tries to predict the conditioned label and, as a result, forces the generator to condition the result on that label.
\textbf{Semantic segmentation}
deals with the classification of the image pixels based on their class labels. For the supervised setting Unet~\cite{unet}, DeepLab~\cite{chen2017deeplab}, DeepLabV3~\cite{chen2017rethinking}, HRNet~\cite{wang2019deep} have shown great performance leaps using a regression loss. For the unsupervised case, more creative solutions are considered. In \cite{chen2019unsupervised,kanezaki2018unsupervised,xia2017w,wang2018unsupervised,croitoru2018unsupervised,sultana2019unsupervised,ji2019invariant,bielski2019emergence} a variety of methods have been used including inpainting, learning feature representation, clustering or video frames comparison. In
\textbf{Clustering,}
 deep learning methods are the current state of the art. JULE~\cite{yang2016joint} and DEPICT~\cite{ghasedi2017deep}, cluster based on a learned feature representation.  IIC~\cite{ji2019invariant} trains a classifier directly.

The most similar approach to ours is FineGAN~\cite{singh2018finegan}, which our generators and discriminators are based upon. However, there are many significant differences and additions: (i) We added a set of encoders, which are trained to support new tasks. (ii) While FineGAN employs one-hot input, our generators use coded input, which is important for our autoencoding path. (iii) We added a skip connection, followed by a mixup module that combines the bypass tensor with the pre-image tensor. The mixup also allows passing only one of the tensors, making either the bypass or the pre-image optional in each flow. (iv) We employ single foreground generator instead of FineGAN's double hierarchical design, where we have found one generator to be dominant and the second one redundant. (v) Our model uses layer normalization~\cite{ba2016layer} instead of batch normalization, which better performs for large number of classes, small batch size, and alternating paths. (vi) We define a new normalization method for the generators, where GLU~\cite{dauphin2017language} activation layers were used as non-linear activations. (vii) We add many losses, regularization terms, and training techniques that were not used in FineGAN, many of which are completely novel, as far as we can ascertain. As a result, our work outperforms FineGAN in all tasks and is capable of performing new tasks that its predecessor could not handle.

Mixup~\cite{mixup} is a technique for applying a weighted sum between two latent variables in order to synthesize a new latent variable. We use it to merge different paths in the model by mixing latent variables that are part coded by the encoder and part produced by the lower levels of the generation. As far as we can ascertain, this is the first usage of mixup to merge information from different paths. Our mixup is applied to image reconstruction in four different locations.

\section{Method}\label{sec:method}

To solve the tasks of clustering, foreground segmentation, and conditional generation with minimal supervision, our method trains multiple neural networks side-by-side. Each task is solved by applying the networks in a task-dependent order. Similarly, the method is trained by applying the networks in two different paths in each iteration, each path is optimized with a specific set of losses. 

\begin{figure}[t]
\centering
\includegraphics[trim={10 450 180 35}, clip, width=\linewidth]{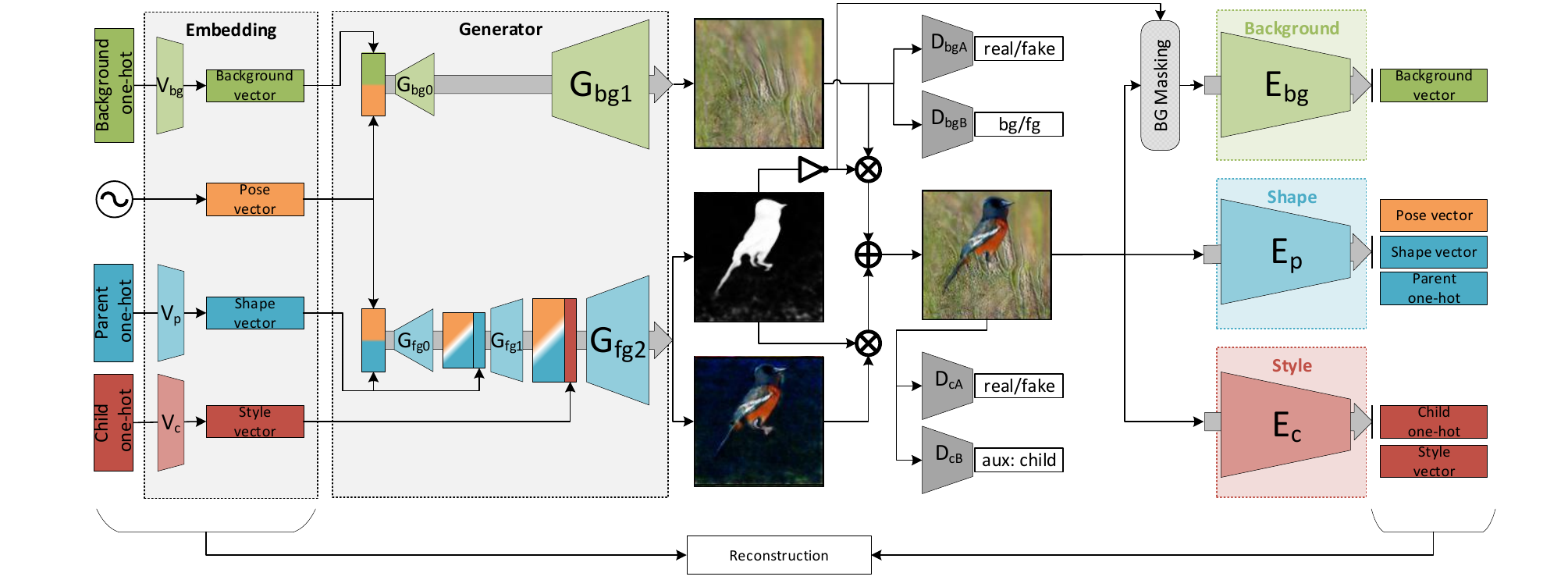}
\caption{\small Flow of the generation path. The generators decode the four priors ($e_{bg},e_p,e_c,z$) and produce three separate images (foreground, background, mask) that are combined into the final image. The generated image is then coded by three encoders to retrieve the latent variables and priors.}
\label{fig:architecture1}
~\\
\centering
\includegraphics[trim={10 65 160 420}, clip, width=1.\linewidth]{images/OneGAN_ECCV_small.pdf}
\vspace{-.20cm}
\caption{\small Flow of the reconstruction path. The same sub-networks are rearranged to perform image reconstruction. The image is coded with the shape and style encoders and then decoded by the foreground generator to produce the foreground image and mask. Then the background encoder and generator code the masked image and produces a background image. The output image combines the foreground and background images. The mixup modules, placed in four different locations, merge the encoders' predicted codes with intermediate stages of the generation, acting as a robust skip connection.
}
\label{fig:architecture2}
\end{figure}

\noindent{\bf Architecture\quad} The architecture of the compound network consists of two generators, three encoders and two discriminators. Fig.~\ref{fig:architecture1} and Fig.~\ref{fig:architecture2} illustrate the two training paths. In the generation path, the generators produce a synthetic image conditioned on selected code, the encoders then retrieve the latent code from the generated images. In the reconstruction path, the encoders code an input image into latent code which is used by the generators to reconstruct the image. The reconstruction path is applied twice in each iteration, once on real images and once on fake images from the generation path. The reconstruction of fake images adds multiple capabilities of self-supervision such as reconstructing the background and mask, which is not applicable with real images without additional supervision. The sub-networks are fully detailed in the supplementary.

\noindent{\bf Generators\quad}
The generation is performed by merging the results of two separate generators that run in parallel to produce the output image. One generator is dedicated for generating the background and the other for the foreground. The generators are conditioned on a two-level hierarchical categorization. Each category has a unique child class $\phi_c$ and a parent class $\phi_p$ shared by multiple child classes. These classes are represented by the one-hot vectors $(e_c, e_p)$. An additional background one-hot vector $e_{bg}$ affects the generation of the background images. Since there is a tight coupling between the class of the object (water bird, tropical bird, etc.) and the expected background, the typology of the background follows the coarse hierarchy, i.e. the parent class. The generator architecture is influenced by.
\begin{equation}
e_{c}[i] = \delta_{i,\phi_c},\quad e_{p}[i] = \delta_{i,\phi_p},\quad e_{bg} = e_p \label{eq:phi}
\end{equation}
The generation starts by converting the one-hot vectors into code vectors using learned embeddings. Such an embedding is often used when working with categorical values.
A fourth vector $z$ is sampled from a multi-variate gaussian distribution to represent non-categorical features.
\begin{equation}
v_{bg} = V_{bg}(e_{bg}),\quad v_{p} = V_{p}(e_p),\quad v_{c} = V_{c}(e_c) \label{eq:V},\quad z \sim \mathcal{N}(0,1)^{d_z}
\end{equation}

The background generator $G_{bg}$ receives the background vector $v_{bg}$ and noise $z$ and produces a background image $I_{bg}$. The foreground generator $G_{fg}$ receives the parent vector $e_p$, child vector $e_c$ and the same $z$ used in the background generation and produces a foreground image $I_{fg}$ and a foreground mask $I_m$. The generator is optimized such that all foreground images with the same $e_p$ will have the same object shape and all images with the same $e_c$ will have a similar object appearance. The latent vector $z$ is implicitly conditioned to represent all non-categorical information, such as pose, orientation, size, etc. It is used in both the background and foreground generation, so that the images produced by both networks will merge into a coherent image.
Each generator is a composition of sub-modules applied back to back, with intermediate pre-images ($A_{bg},A_{fg}$):
\begin{align}
     I_{bg} &= G_{bg_1}(A_{bg}),\quad 
     &A_{bg} &= G_{bg_0}(v_{bg}, z)
     \label{eq:Gbg} \\
     (I_{fg}, I_m) &= G_{fg_2}(G_{fg_1}(A_{fg}, v_p), v_c),\quad
     &A_{fg} &= G_{fg_0}(v_p, z)
    \label{eq:Gfg}
\end{align}

The final generated image is: (where $\circ$ denotes element-wise multiplication)
\begin{equation}
I = I_{bg} \circ (1 - I_{m}) + I_{fg} \circ I_{m} \label{eq:G}
\end{equation}

\noindent{\bf Encoders\quad} 
Unlike FineGAN, which performs only the generation task, our method requires the use of encoders. We introduce three encoders: background encoder $E_{bg}$, shape encoder $E_{p}$, and style encoder $E_{c}$. They run in semi-parallel to predict both the latent codes ($v_{bg},v_p,v_c, z$) of an input image and the underlying one-hot vectors ($e_{bg},e_p,e_c$). All encoders are fed with image $I$ as input. The background encoder is also fed with the mask $I_m$. During image reconstruction, there is no initial image mask, therefore it first has to be generated by encoding the shape and style features and applying the foreground generator. The lack of ground-truth mask is why the encoders do not run fully in parallel. In addition, the background and shape encoders also produce bypass tensors ($B_{bg},B_{fg}$) to be used as skip connections between the encoders and the generators.
\begin{align}
    (B_{bg}) &= E_{bg}(I, I_m) \label{eq:Ebg} \\
    (\hat{e}_p ,\mu_p, \sigma_p, B_{fg}, \mu_z, \sigma_z) &= E_{p}(I) \label{eq:Ep}\\        
    (\hat{e}_c ,\mu_c, \sigma_c) &= E_{c}(I) \label{eq:Ec}
\end{align}
%
Following Variational Auto-Encoder literature, ($\mu, \sigma$) are three paired vectors of sizes ($d_{z}, d_{p}, d_{c}$) defining the mean and variance to sample each element of ($\hat{z}, \hat{v}_{p}, \hat{v}_{c}$) from a gaussian distribution .

\noindent{\bf Mixup\quad} 
At the intersection between the encoders and generators, we introduce a novel method to merge information coded by the encoders and information produced by the embeddings and lower levels of the generators. The mixup module~\cite{mixup}, mixes two input variables with a weight parameter $\beta$. The rationale behind this application is that during generation there is no data coming from the encoders, so the mixup is turned off and only information from the embeddings and lower levels of the generators are passed forward. During reconstruction, we want our method to utilize the skip connections to improve performance and also use the predicted embeddings $(v_p,v_c)$ to represent the object's shape and style. The contrast between the two paths leads to a difficulty in optimizing them simultaneously. The introduction of the mixup simplifies this by having both paths active during forward path and back-propagation. In contrast to regular residual connections, the ever changing $\beta$ used in the mixup forces both inputs to be independent representations and not complement each other. 

The mixup modules at the vector embeddings level (mixup1 and mixup2 in Fig.~\ref{fig:architecture2}) mix the vectors ($v_p, v_c$) given by the embeddings ($V_p, V_c$), Eq.~\ref{eq:V}, with the predicted vectors ($\hat{v}_p, \hat{v}_c$) produced by the encoders ($E_p, E_c$), Eq.~\ref{eq:Ep},\ref{eq:Ec}. The mixture of features leads to both the embeddings and the encoders being optimized for reconstructing the object. This has two benefits. First, it trains the encoders to properly code the images, which improves clustering and learns image-to-image translation implicitly. Second, it trains the embeddings to represent the real object classes, which improves the generation task.

The mixup modules at the skip connections (mixup3 and mixup4 in Fig.~\ref{fig:architecture2}) mix the pre-image tensors ($A_{bg}, A_{fg}$), Eq.~\ref{eq:Gbg},\ref{eq:Gfg}, with the bypass tensors ($B_{bg},B_{fg}$), Eq.~\ref{eq:Ebg},\ref{eq:Ep}. It serves to create the condition where the reconstruction path will be simultaneously dependent on the bypass and on the lower stage of the generators. This way, at any time we can choose any $\beta$ or even pass only the bypass or only the pre-image and result in an almost identical image.

Given two inputs and a parameter $\beta$, the mixup is defined as follows: 
\begin{equation}\label{eq:mixup}
\centering
\begin{split}
v_{p_{mix}} = v_p \circ (1 - \beta_1) + \hat{v}_p \circ \beta_1,&\quad v_{c_{mix}} = v_c \circ (1 - \beta_2) + \hat{v}_c \circ \beta_2 \\
A_{fg_{mix}} = A_{fg} \circ (1 - \beta_3) + B_{fg} \circ \beta_3,&\quad A_{bg_{mix}} = A_{bg} \circ (1 - \beta_4) + B_{bg} \circ \beta_4
\end{split}
\end{equation}
In our implementation, $\beta_1,\beta_2 \in [0,1]$ and $\beta_3,\beta_4 \in [0.5,1]$, are sampled in each iteration for each instance in the batch. At reconstruction, the mixed features ($v_{p_{mix}}, v_{c_{mix}}, A_{fg_{mix}}, A_{bg_{mix}}$) replace the features ($v_{p}, v_{c}, A_{fg}, A_{bg}$) in Eq.~\ref{eq:Gbg},\ref{eq:Gfg} as input to the generators. For illustration, please refer to Fig.~\ref{fig:architecture2}.

\noindent{\bf GLU Layer Normalization\quad} 
Following StackGANv2 and FineGAN architecture, we apply GLU~\cite{dauphin2017language} activation in the generators. Due to the multiple paths, the large scale and high complexity of our method, batch normalization was unstable for our low batch size, and, increasing the batch size was not an option. As a solution, we switched to layer normalization, which is not affected by the batch size. We fused the normalization and activation into a single module termed ``GLU Layer Normalization''. Given an input $x$ with $x_L,x_R$ representing an equal split in the channel axis (left/right): 
\begin{equation}
\begin{split}
    \text{GLU}(x_L,x_R) &= x_L \circ \text{Sigmoid}(x_R) \\
    \text{GLU-LNorm}(x_L,x_R) &= \text{GLU}(\text{LNorm}(x_L),x_R)
\end{split}
\end{equation}
In this method, the normalization is only applied on $x_L$. The input to the sigmoid, $x_R$, is not normalized. This is favorable, because $x_R$ serves as a mask on $x_L$, and normalizing it across the channels contradicts this goal.

\noindent{\bf Discriminators\quad} 
Following FineGAN, the two discriminators are adversarial opponents on the outputs $I_{bg}, I$. 
The background discriminator $D_{bg}$ has two tasks, with a separate output for each. The tasks are as follows: (i) patch-wise prediction if the input image is real or fake when presented with either a real or fake background image, annotated as $D_{bg_A}$. (ii) patch-wise prediction if the input image is a background image or not when presented with either a real background image or a real object image, annotated as $D_{bg_B}$. The background generator is hereby optimized to generate images that look like real images and do not contain object features. In addition, when performing the reconstruction path on fake images, we also extract a hidden layer output and apply perceptual loss between generated and reconstructed backgrounds, annotated as $D_{bg_C}$, to reduce the perceptual distance between the original and the reconstructed image. 

The image discriminator $D_c$ receives real images from $X_c$ or generated fake images, and also has two tasks: (i) predict if the input image is real or fake, annotated as $D_{c_A}$. (ii) predict the child class $\phi_c$ of the image, annotated as $D_{c_B}$, as in all InfoGAN-influenced methods. This trains the foreground generator to generate images that look real and represent the conditioned child class. In addition, we also extract a hidden layer output and apply perceptual loss between generated and reconstructed foreground images, annotated as $D_{c_C}$.

\section{Training}\label{sec:training}
To train to solve various tasks, we perform in each iteration two different paths through the model, by connecting the various sub-networks in a specific order. 


\subsection{Generation path}\label{sec:generation_path}
\begin{figure*}[t]
\centering
\includegraphics[width = 0.32\textwidth]{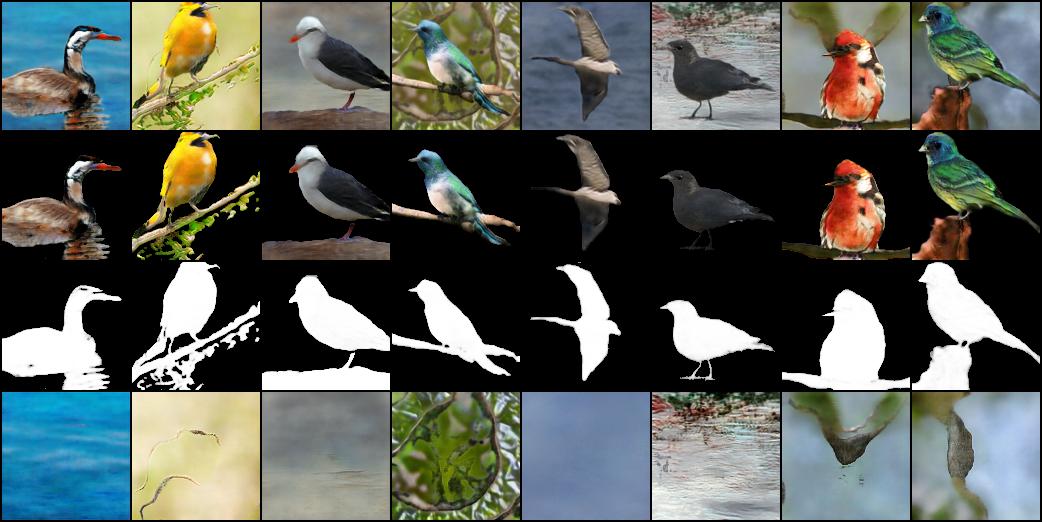}
\includegraphics[width = 0.32\textwidth]{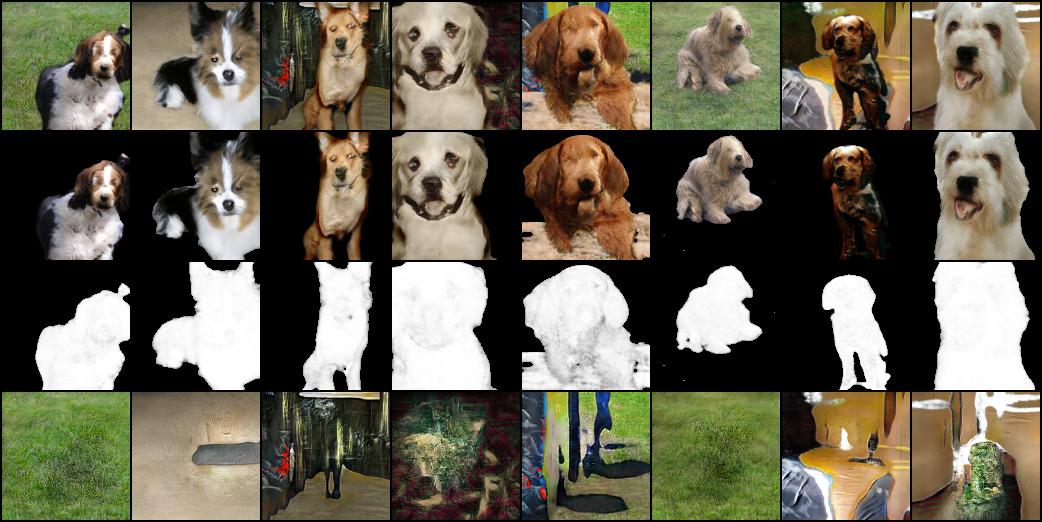}
\includegraphics[width = 0.32\textwidth]{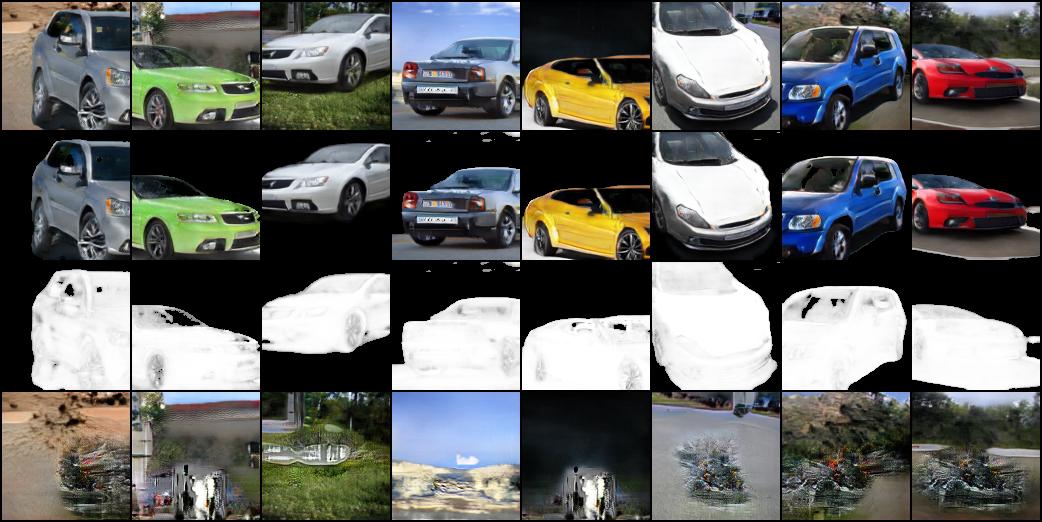} \\
\vspace{-.35cm}
\caption{Image Generation for each dataset. From top to bottom: (i) final image, (ii) foreground, (iii) foreground mask, (iv) background.}
\label{fig:generation}
\vspace{-.35cm}
\end{figure*}

The generation path is described in Fig.~\ref{fig:architecture1}, Eq.~\ref{eq:V}--\ref{eq:G}. For illustrations, see Fig.~\ref{fig:generation}. The inputs for this path are $e_{bg},e_{p},e_{c},z$. During generation, the model learns to generate image $I$ in a way that relies on generating a background $I_{bg}$, foreground $I_{fg}$, and mask $I_m$ images. The discriminators are trained along with the generators and produce an adversarial training signal. In addition, the encoders are also trained to retrieve the latent variables from the generated images, as a self-supervised task.

The losses in this path can be put into four groups: adversarial losses, classification losses, distance losses, and regularizations. For brevity, $e$ represents the dependence on all prior codes ($e_{bg}, e_p, e_c$). Similarly, $G(e,z)$ represents the full generation of the final image, Eq.~\ref{eq:Gbg}--\ref{eq:G}.

\noindent{\bf Adversarial losses\quad}
These involve the two discriminators and are derived from the minimax equation:
$\min_G \max_D \E_x[\log(D(x))] + \E_z[\log(1 - D(G(z)))]$, for a generic generator $G$ and discriminator $D$. The concrete GAN loss is the sum of the losses for the separation between real/fake background, the separation between background/object and the separation between real/fake object. 

For the discriminators, where $X_{bg}, X_c$ are the sets of real background images and real object images, the losses are:
\begin{equation}
\begin{split}
\mathcal{L}_{D_{bg_A}} = &\E_{x\sim X_{bg}}[\log(D_{bg_A}(x))] + \E_{e_{bg},z}[\log(1 - D_{bg_A}(G_{bg}(e_{bg}, z)))] \\
\mathcal{L}_{D_{bg_B}} = &\E_{x\sim X_{bg}}[\log(D_{bg_B}(x))] + \E_{x\sim X_c}[\log(1 - D_{bg_B}(x))] \\
\mathcal{L}_{D_{c_A}} = &\E_{x\sim X_c}[\log(D_{c_A}(x))] + \E_{e,z}[\log(1 - D_{c_A}(G(e, z)))] \\
\mathcal{L}_D = & 10 \cdot \mathcal{L}_{D_{bg_A}} + \mathcal{L}_{D_{bg_B}} + \mathcal{L}_{D_{c_A}}
\end{split}
\end{equation}

For the generators, the losses are:
\begin{equation}
\begin{split}
    \mathcal{L}_{G_{bg_A}} &= \E_{e_{bg},z}[\log(D_{bg_A}(G_{bg}(e_{bg}, z)))] \\
    \mathcal{L}_{G_{bg_B}} &= \E_{e_{bg},z}[\log(D_{bg_B}(G_{bg}(e_{bg}, z)))] \\
    \mathcal{L}_{G_{c_A}} &=  \E_{e,z}[\log(D_{c_A}(G(e, z)))] \\
    \mathcal{L}_G &=  10 \cdot \mathcal{L}_{G_{bg_A}} + \mathcal{L}_{G_{bg_B}} + \mathcal{L}_{G_{c_A}}
\end{split}
\end{equation}

\noindent{\bf Classification losses\quad}
These losses optimize the generators to generate distinguished images for each style and shape priors and optimize the encoders to retrieve the prior classes. We use the cross entropy loss between the conditioned classes ($\phi_p, \phi_c$) and the encoders' predictions ($\hat{e}_p, \hat{e}_c$) form Eq.\ref{eq:Ep},\ref{eq:Ec}. In addition, we use the auxiliary task $D_{C_B}$.
\begin{equation}
\mathcal{L}_E =  \text{CE}(\hat{e}_p, \phi_p) + \text{CE}(\hat{e}_c, \phi_c) + \text{CE}(D_{c_B}(I), \phi_c)
\end{equation}

\noindent{\bf Distance losses\quad}
We train the encoders to minimize the mean squared error between the vectors in the latent space produced during generation and their predicted counterparts. 
These vectors are used in the reconstruction path, thus this self-supervised task assists in this regard. We minimize the distance between the pre-images and bypasses ($A_{bg}, A_{fg}, B_{bg}, B_{fg}$), and between the latent vectors ($v_p, v_c$) and the mean vectors $\mu_p,\mu_c$ used to sample the latent code ($\hat{v}_p, \hat{v}_c$).
\begin{equation}
\mathcal{L}_{\text{MSE}} = \text{MSE}(v_c, \mu_c) + \text{MSE}(v_p, \mu_p) + \text{MSE}(A_{fg}, B_{fg}) + \text{MSE}(A_{bg}, B_{bg})
\end{equation}

\noindent{\bf Regularization losses\quad}
For regularization, a loss term is applied on the latent codes ($v_{bg}, v_p, v_c$), annotated as $\mathcal{L}_{R_v}$, and on the foreground mask $I_m$, annotated as $\mathcal{L}_{R_M}$. They are detailed in the supplementary.

All the losses are summed together to the total loss:
\begin{equation}\label{eq:lossesGEN}
\mathcal{L}_{\text{GEN}} = \mathcal{L}_G + \mathcal{L}_E + \mathcal{L}_{\text{MSE}} + 0.1 \cdot \mathcal{L}_{R_v} + 2 \cdot \mathcal{L}_{R_M}
\end{equation}

\subsection{Reconstruction path}\label{sec:reconstruction_path}
The reconstruction path is described in Fig.~\ref{fig:architecture2}. For illustrations, see Fig.~\ref{fig:reconstruction}. The input is an image $I$. The precise flow is: (1) encode the foreground through the shape and style encoders ($E_p, E_c$), Eq.~\ref{eq:Ep},\ref{eq:Ec}, (2) generate a foreground image and mask with the foreground generator ($G_{fg}$), Eq.~\ref{eq:Gfg}, (3) encode the image and mask through the background encoder ($E_{bg}$), Eq.~\ref{eq:Ebg}, (4) generate the background image with the background generator ($G_{bg}$), Eq.~\ref{eq:Gbg}, and (5) compose the final image with $I_{fg}, I_{bg}, I_m$, Eq.~\ref{eq:G}. In addition, the mixup is applied as in Eq.~\ref{eq:mixup} between encoding and generation. This path optimizes the clustering and segmentation tasks directly and also implicitly optimizes the generation task by reconstruction real images. We perform the reconstruction path on both real images and generated images from the generation path. This fully utilizes the information available to learn all tasks with minimal supervision.



The losses in this path can be put in three groups: statistical losses, reconstruction losses, and perceptual losses.

\noindent{\bf Statistical losses\quad}
As in Variational Auto-Encoders~\cite{vae}, we compare the Kullback-Leiber Divergence between the latent variables encoded by the encoders ($\hat{v}_p, \hat{v}_c, \hat{z}$) to a multivariate gaussian distribution.
For the pose vector $z$, we used the standard normal distribution with covariance matrix equal to the identity matrix ($\Sigma = I_{d_z}$) and a zero mean vector ($\mu = \textbf{0}$). For the shape and style vectors ($\hat{v}_p, \hat{v}_c$) we still use identity $\Sigma$, but since they should match their latent code ($v_p, v_c$), we use these latent codes as the target mean.
\begin{small}
\begin{equation}
\begin{split}
\mathcal{L}_{\text{VAE}_p} &= \KL (\mathcal{N}(\mu_{p}, \text{diag}(\sigma_{p})) \Vert \mathcal{N}(v_p, I_{d_p})),   \mathcal{L}_{\text{VAE}_c} = \KL (\mathcal{N}(\mu_{c}, \text{diag}(\sigma_{c})) \Vert \mathcal{N}(v_c, I_{d_c}))  \\
\mathcal{L}_{\text{VAE}_z} &= \KL (\mathcal{N}(\mu_{z}, \text{diag}(\sigma_{z})) \Vert \mathcal{N}(\textbf{0},~~~ I_{d_z})),  \mathcal{L}_{\text{VAE}} = \mathcal{L}_{\text{VAE}_p} + \mathcal{L}_{\text{VAE}_c} + \mathcal{L}_{\text{VAE}_z}
\end{split}
\raisetag{12pt}
\end{equation}
\end{small}

\noindent{\bf Reconstruction losses\quad}
The reconstruction losses are a set of L1 losses that compare the difference between the input image to the output.
The network trains at reconstructing both real and fake images. For fake images, we have the extra self-supervision to also compare reconstruction of the background image and foreground mask.
\begin{equation}
\mathcal{L}_{\text{REC}} = 
    \begin{cases}
    \text{L1}(I, \hat{I}) &, \text{real}\\
    \text{L1}(I, \hat{I}) + \text{L1}(I_{bg}, \hat{I}_{bg}) + \text{L1}(I_m, \hat{I}_m) &, \text{fake}
    \end{cases}
\end{equation}

\noindent{\bf Perceptual losses}\label{par:perceptual_loss}
Comparing images to their ground-truth counterpart is known to produce blurred images; Perceptual loss~\cite{johnson2016perceptual} is known to aid in producing sharper images with more visible context~\cite{zhang2018unreasonable} by comparing the images on the feature level as well. The perceptual loss is often used along with a pre-trained network, but this relies on added supervision. In our case, we use the discriminators as feature extractors. We use the notation $D_{bg_C}, D_{c_C}$ from Sec.~\ref{sec:method} to describe the extraction of the hidden layers used for this comparison.\\
\begin{equation}
\mathcal{L}_{\text{PER}} = 
    \begin{cases}
    \text{L2}_{D_{c_C}}(I, \hat{I}) & , \text{real}\\
    \text{L2}_{D_{c_C}}(I, \hat{I}) + \| D(D_{bg_C})- D(\hat{I}_{bg})\|^2 
    & , \text{fake}
    \end{cases}
\end{equation}
%

All the losses are summed together to the total loss:
\begin{equation}
\mathcal{L}_{\text{AE}} = \mathcal{L}_{\text{GEN}} + \mathcal{L}_{\text{VAE}} + \mathcal{L}_{\text{REC}} + \mathcal{L}_{\text{PER}}
\end{equation}

\subsection{Multi-phase training}\label{sec:multiphase}
In order to simplify training, instead of training both paths at once, we schedule the training process by phases. The phases are designed to train the network for a gradually increasing subset of tasks, starting from image-level tasks (generating images) to semantic tasks (semantic segmentation of the foreground, and semantic clustering) that benefit from the capabilities obtained in the generation path.
In the first phase we only perform the generation path~\ref{sec:generation_path} and in the second phase we add the reconstruction path~\ref{sec:reconstruction_path}.

Without multi-phase training, the networks would be trained to generate and reconstruct images simultaneously. While the generation flow encourages a separation between the background and foreground components, the reconstruction flow resists this separation due to the trivial solution of encoding and decoding the image in one of the paths (foreground or background) and applying an all-zero or all-one mask. 
In the experiments, in Tab.~\ref{table:generation},\ref{table:segmentation}, we show that without multi-phase the model is incapable of learning any task.

In this controlled environment, the generators are much more likely to converge to the required setting. After a decent amount of iterations, determined in advance by a hyper-parameter, the second phase kicks in, where the model is also trained to reconstruct images, which will train the encoders on top of the generator instead of breaking it.

When entering Phase II, the fake images for both discriminators can be a result of either (i) generation path, (ii) fake image reconstruction, or (iii) real image reconstruction. We noticed that images from the reconstruction paths fail to converge to real-looking images when the discriminators were only trained by the generation path outputs. We hypothesized that this is probably due to each path producing images from a different source domain and these paths can generate very different images during training and the discriminators get overwhelmed by the different tasks and are not able to optimize them simultaneously. To solve this, upon entering Phase II, we clone each discriminator ($D_c, D_{bg}$) twice and associate one separate clone for each path, resulting in a total of three background discriminators and another three for the foreground. In this setting, each path receives the adversarial signal that is concentrated only at improving that path.

\section{Experiments}\label{sec:experiments}

\begin{figure}[t]
\centering
\includegraphics[width = 0.32\linewidth,trim={0 390 0 0}, clip]{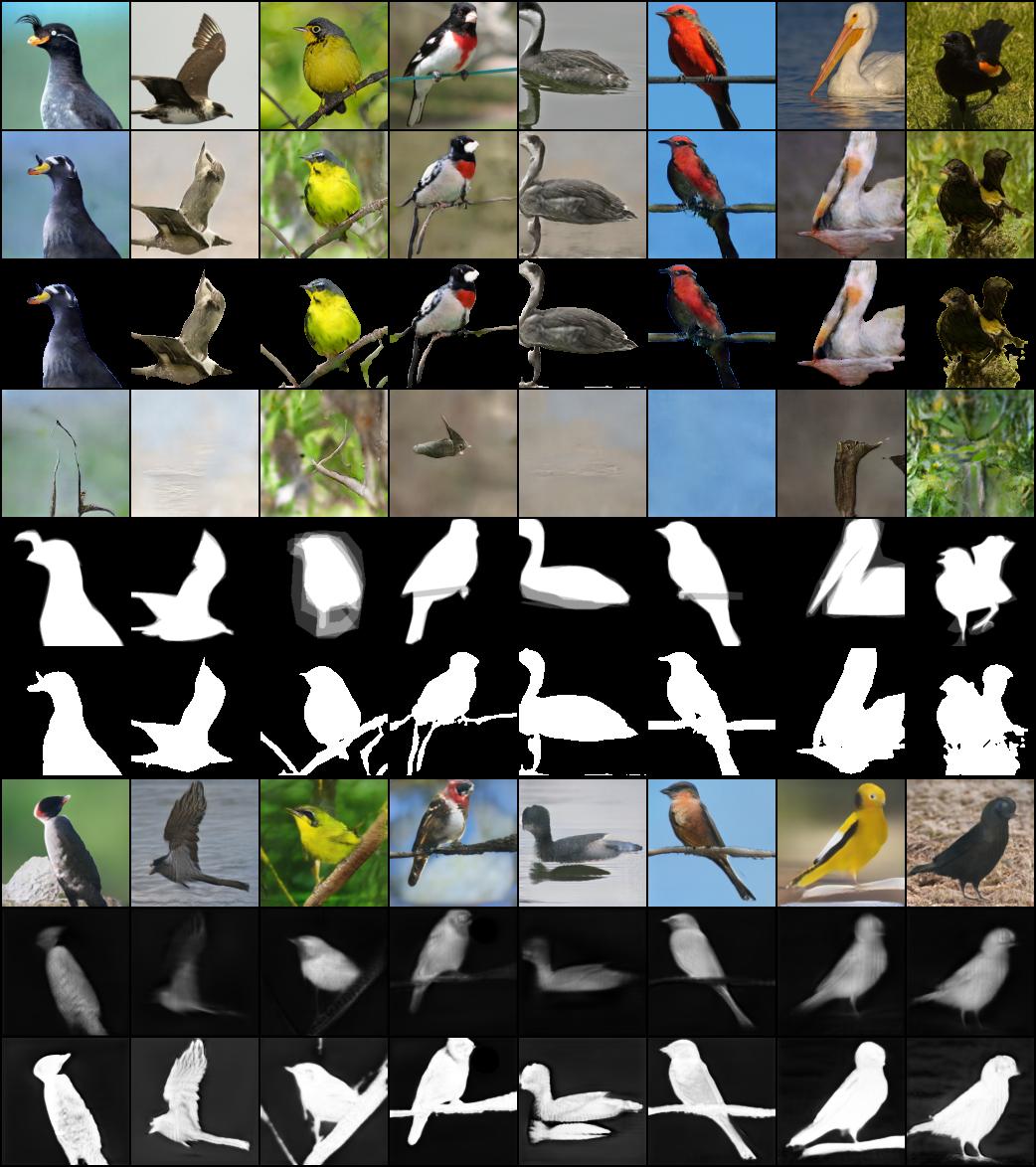}
\includegraphics[width = 0.32\linewidth,trim={0 390 0 0}, clip]{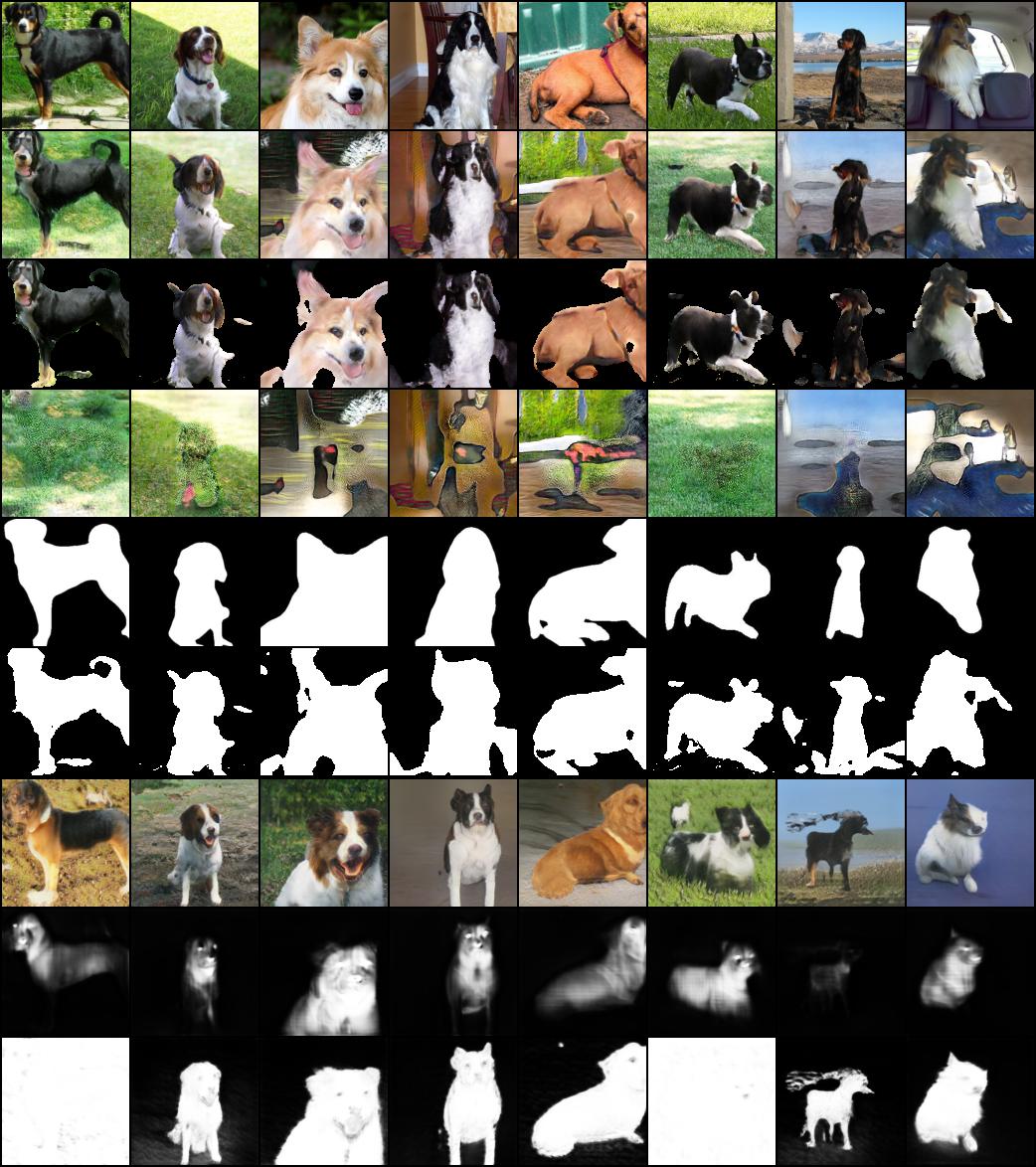} 
\includegraphics[width = 0.32\linewidth,trim={0 390 0 0}, clip]{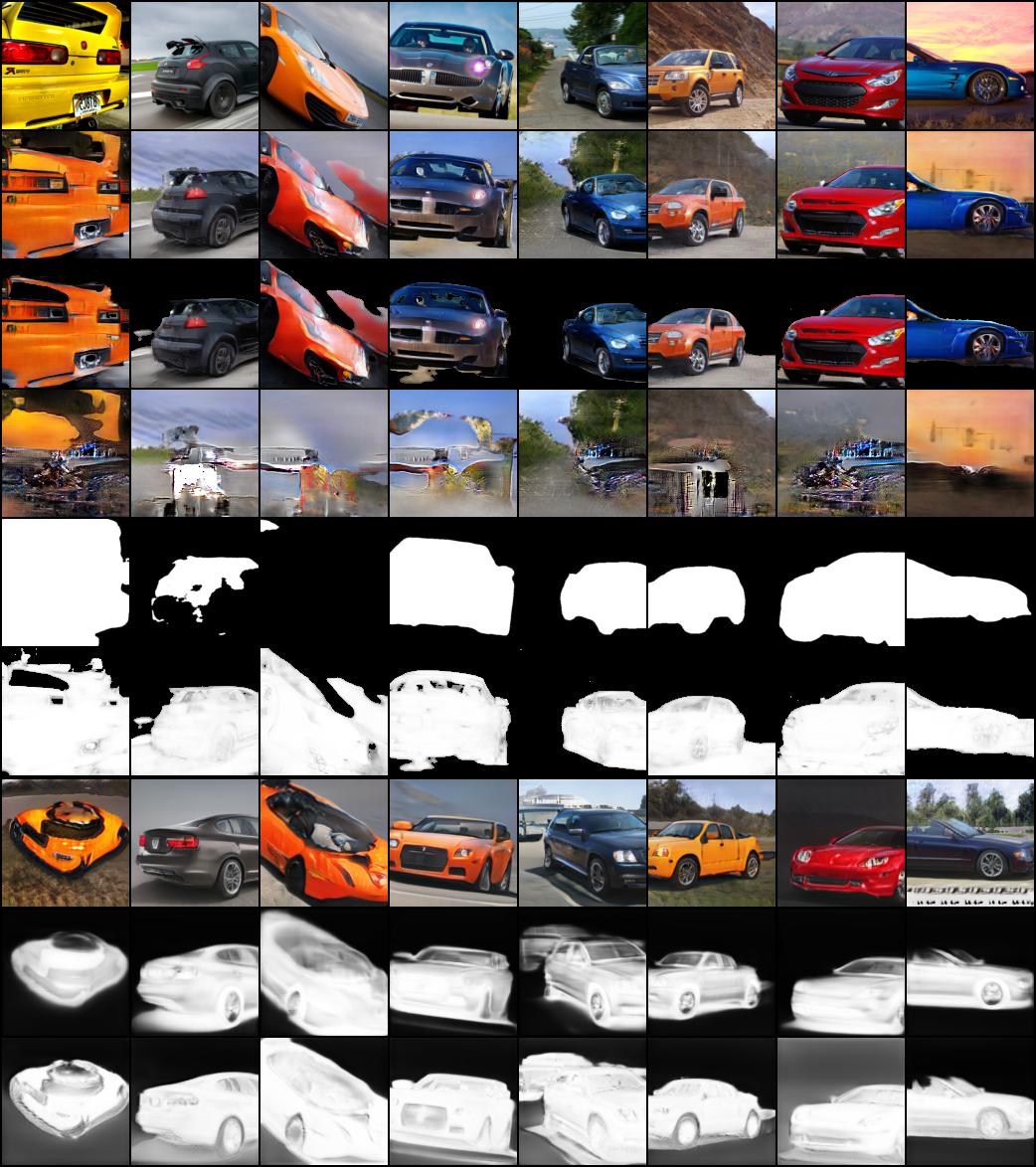} \\
\vspace{-.35cm}
\caption{Image reconstruction for each dataset. From top to bottom: (i) real image, (ii) reconstructed image, (iii) reconstructed foreground, (iv) reconstructed background, (v) ground-truth foreground mask, (vi) predicted foreground mask.}
\label{fig:reconstruction}
\vspace{-.35cm}
\end{figure}

\begin{figure*}[t]
\centering
\begin{minipage}{0.485\textwidth}
\includegraphics[width=\textwidth]{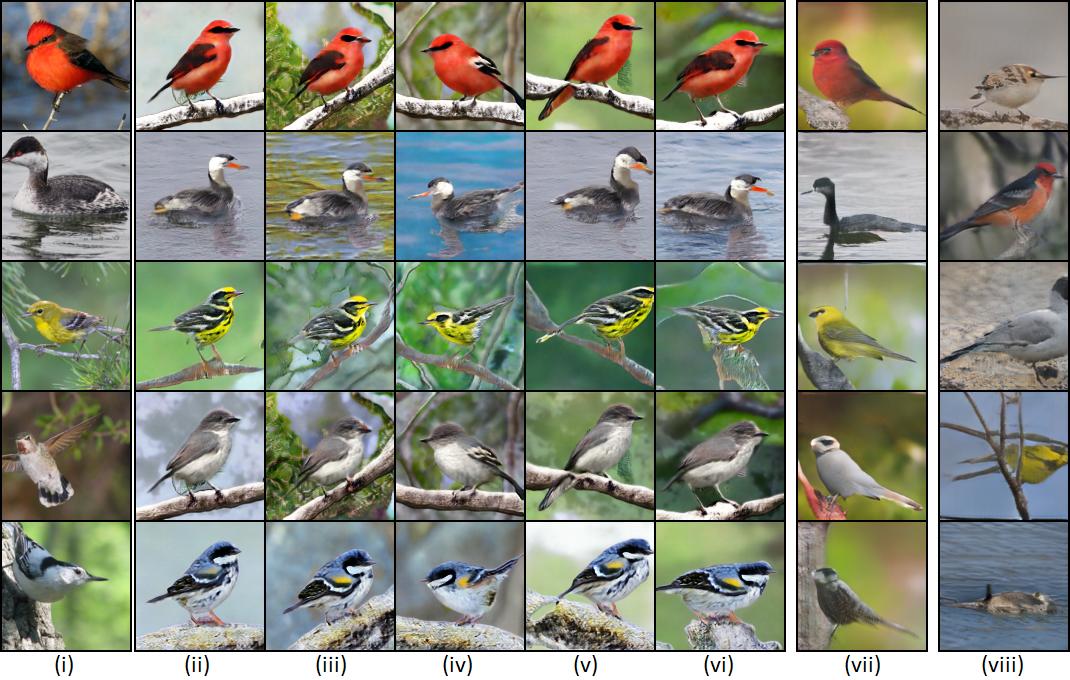}
\vspace{-.35cm}
\caption{Conditional Generation. From left to right: (i) real images, (ii-vi) generation of images with the encoded parent and child codes and a different vector $z$ per column, (vii) FineGAN~\cite{singh2018finegan} + our encoders, (viii) StackGANv2~\cite{zhang2018stackgan++} + our encoders.
\label{fig:conditional}}
\end{minipage} \hfill%
\begin{minipage}{0.485\textwidth}
\includegraphics[width=\textwidth]{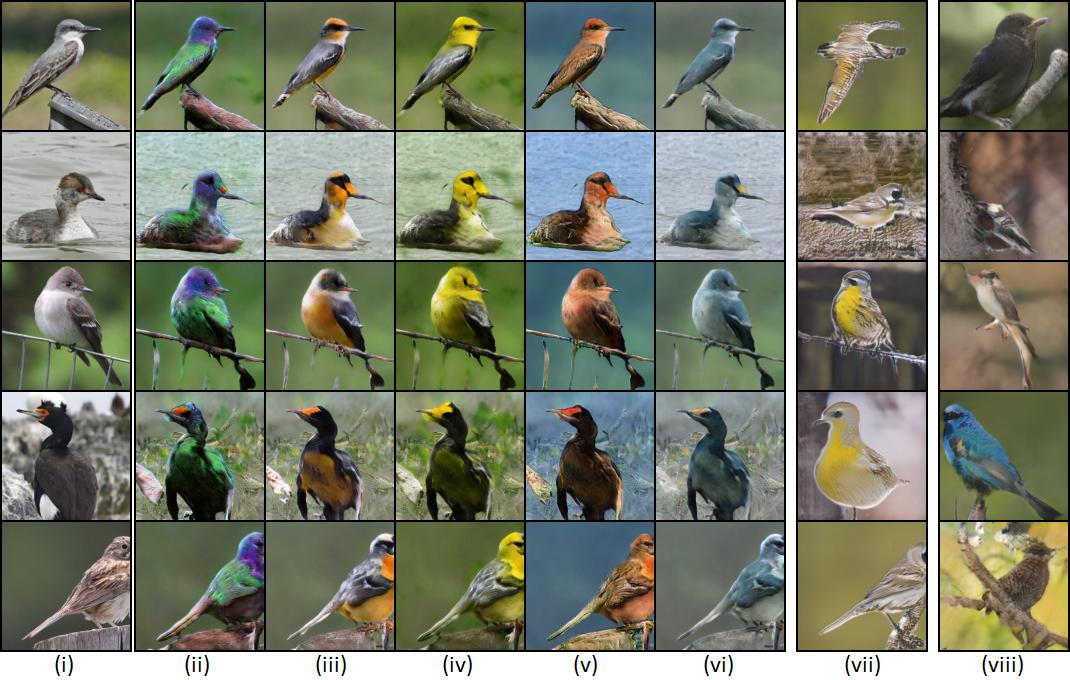}
\vspace{-.35cm}
\caption{Style Transfer. From left to right: (i) real images. (ii-vi) reconstructed images when the child code $e_c$ is switched with a code from a selected category, (vii) FineGAN~\cite{singh2018finegan} + our encoders, (viii) StackGANv2~\cite{zhang2018stackgan++} + our encoders.
\label{fig:transfer}
}
\end{minipage}
\vspace{-.20cm}
\end{figure*}

\begin{table}[t]
\centering
\caption{Quantitative generation results. FID$\downarrow$, IS$\uparrow$, CFID$\downarrow$, CIS$\uparrow$}
\vspace{-.20cm}
\label{table:generation}
\begin{tabular}{@{}l@{}c@{~}c@{~}c@{~}cc@{~}c@{~}c@{~}cc@{~}c@{~}c@{~}c}
\toprule
    & \multicolumn{4}{c}{Birds} & \multicolumn{4}{c}{Dogs} & \multicolumn{4}{c}{Cars} \\ 
\cmidrule(lr){2-5}
\cmidrule(lr){6-9}
\cmidrule(lr){10-13}
Model           & FID       & IS        & CFID      & CIS       & FID        & IS       & CFID      & CIS       & FID        & IS       & CFID      & CIS       \\ 
\midrule
Dataset         & 0         & 163.6     & 0         & 47.9      & 0         & 114.2     & 0         & 77.1      & 0         & 163.1     & 0         & 55.4      \\ 
\midrule
StackGANv2      & 21.4      & 67.0      & 96.8      & 15.0      & 56.7      & 82.4      & 184.7     & 10.2      & 25.0      & 88.1      & 190.3     & 13.3      \\
FineGAN         & 23.0      & 66.4      & 65.3      & 24.7      & 54.9      & 83.1      & 100.4      & 15.7      & 24.8      & 86.2      & 126.0      & 13.6      \\ 
\midrule
OneGAN          & {\bf20.5} & {\bf67.4} & {\bf55.2} & {\bf30.7} & {\bf48.7} & {\bf89.7} & {\bf92.0} & {\bf19.6} & {\bf24.2} & {\bf90.3} & {\bf100.7} & {\bf18.7} \\ 
no real recon   & 22.3      & 65.6      & 58.6      & 25.6      & 55.4      & 84.2      & 95.3      & 17.0      & 25.1      & 88.2      & 104.3      & 15.5      \\
Phase I only    & 23.9      & 63.2      & 59.1      & 21.6      & 56.1      & 82.0      & 97.8      & 16.8      & 25.4      & 87.7      & 106.1      & 13.4      \\
no multi-phase  & 196.5     & 11.0      & 356.1     & 2.3       & 217.8     & 16.9      & 543.2     & 1.7       & 264.7     & 23.4      & 767.9      & 3.9        \\ 
\bottomrule
\end{tabular}
\vspace{-.60cm}
\end{table}

\begin{table}[t]
\centering
\centering
\caption{Segmentation and clustering results. $^\mathsection$unfair upper bound results, obtained by selecting the best result out of many. $^\dag$provided by \cite{singh2018finegan}. $^*$model performed task by using our encoders. \xmark model cannot perform task. Higher is better in all scores.}
\label{table:segmentation}\label{table:clustering}
\vspace{-.20cm}
\begin{tabular}{@{}l@{~}c@{~}c@{~}c@{~}c@{~}c@{~}cc@{~}c@{~}c@{~}c@{~}c@{~}c@{}}
\toprule
    & \multicolumn{6}{c}{Segmentation} & \multicolumn{6}{c}{Clustering} \\
\cmidrule(lr){2-7}
\cmidrule(lr){8-13}
                    & \multicolumn{2}{c}{Birds} & \multicolumn{2}{c}{Dogs} &         \multicolumn{2}{c}{Cars} & \multicolumn{2}{c}{Birds} & \multicolumn{2}{c}{Dogs} & \multicolumn{2}{c}{Cars} \\
\cmidrule(lr){2-3}
\cmidrule(lr){4-5}
\cmidrule(lr){6-7}
\cmidrule(lr){8-9}
\cmidrule(lr){10-11}
\cmidrule(lr){12-13}
Model                    & \multicolumn{1}{@{}c@{}}{IOU} & \multicolumn{1}{@{}c}{DICE}       
                         & \multicolumn{1}{@{}c@{}}{IOU} & \multicolumn{1}{@{}c}{DICE} 
                         & \multicolumn{1}{@{}c@{}}{IOU} & \multicolumn{1}{@{}c}{DICE} 
                                                                                                        & ACC       & NMI       & ACC       & NMI       & ACC       & NMI       \\ 
\midrule
ReDO                            & 46.5      & 60.2      & 38.4      & 52.8      & 16.2      & 26.2      & \xmark    & \xmark    & \xmark    & \xmark    & \xmark    & \xmark    \\
WNet                            & 24.8      & 38.9      & 47.7      & 62.1      & 52.8      & 67.6      & \xmark    & \xmark    & \xmark    & \xmark    & \xmark    & \xmark    \\
UISB$^\mathsection$             & 44.2      & 60.1      & 62.7      & 75.5      & 64.7      & 77.5      & \xmark    & \xmark    & \xmark    & \xmark    & \xmark    & \xmark    \\
IIC-seg stf-3$^\mathsection$   & 36.5      & 50.2      & 58.5      & 71.5      & 58.5      & 71.5      & \xmark    & \xmark    & \xmark    & \xmark    & \xmark    & \xmark    \\
IIC-seg stf$^\mathsection$    & 35.2      & 50.4      & 56.6      & 70.2      & 58.8      & 71.7      & \xmark    & \xmark    & \xmark    & \xmark    & \xmark    & \xmark    \\ 
\midrule
JULE$^\dag$                     & \xmark    & \xmark    & \xmark    & \xmark    & \xmark    & \xmark    & .045      & .204      & .043      & .142      & .046      & .232      \\
DEPICT$^\dag$                   & \xmark    & \xmark    & \xmark    & \xmark    & \xmark    & \xmark    & .061      & .290      & .052      & .182      & {\bf.063} & {\bf.329} \\
IIC-cluster                     & \xmark    & \xmark    & \xmark    & \xmark    & \xmark    & \xmark    & .084      & .345      & .060      & .200      &  .056     & .254      \\
StackGANv2                      & \xmark    & \xmark    & \xmark    & \xmark    & \xmark    & \xmark    & .057$^*$  & .253$^*$  & .040$^*$  & .139$^*$  & .039$^*$  & .174$^*$  \\
FineGAN                         & 44.5$^*$  & 56.9$^*$  & 48.7$^*$  & 59.3$^*$  & 53.2$^*$  & 60.3$^*$  & .086$^*$  & .349$^*$  & .059$^*$  & .194$^*$  & .051$^*$  & .233$^*$  \\ 
\midrule
OneGAN                          & {\bf55.5} & {\bf69.2} & {\bf71.0} & {\bf81.7} & {\bf71.2} & {\bf82.6} & {\bf.101} & {\bf.391} & {\bf.073} & {\bf.211} & .060      & .272      \\
no real recon                  & 53.5      & 67.7      & 67.1      & 78.6      & 69.8      & 81.1      & .095      & .389      & .062      & .194      & .057      & .250      \\
Phase I only                    & 45.7      & 60.6      & 65.1      & 77.3      & 64.8      & 75.9      & .084      & .352      & .058      & .175      & .052      & .244      \\
no~multi-phase                  & 28.2      & 43.2      & 7.4       & 13.6      & 45.9      & 60.5      & .050      & .216      & .019      & .082      & .041      & .208      \\ 
\bottomrule
\end{tabular}
\vspace{-.20cm}
\end{table}


\begin{table*}[t]
\centering
\caption{Ablation studies on CUB: (a) normalization methods, (b) modules' behaviour, and (c) losses. Measuring FID and C-IS for generation and IOU for segmentation.}
\label{table:ablation}
\begin{adjustbox}{max width=.999\textwidth}
\begin{tabular}{ccc}
\begin{tabular}{l@{~}c@{~}c@{~}c@{~}c}
\toprule
\multicolumn{1}{l}{Model}   & FID & C-IS & IOU  \\
\midrule
OneGAN      & 20.5 & 30.7 & 55.5   \\
GLU-INorm   & 122.0 & 10.2 & 31.3   \\
LNorm       & 87.5 & 14.5 & 45.4   \\
INorm       & 103.4 & 9.8 & 30.1   \\
\bottomrule
\end{tabular}&
\begin{tabular}{l@{~}c@{~}c@{~}c@{~}c}
\toprule
\multicolumn{1}{l}{Model}  & FID & C-IS & IOU  \\
\midrule
OneGAN          & 20.5 & 30.7 & 55.5   \\
no bypass       & 21.2 & 22.8 & 53.3   \\
no mixup(1,2)   & 22.6 & 17.5 & 54.1   \\
no mixup(3,4)   & 20.9 & 22.2 & 53.8   \\
\bottomrule
\end{tabular}&
\begin{tabular}{l@{~}l@{~}c@{~}c@{~}c}
\toprule
\multicolumn{1}{l}{Model}   & FID & C-IS & IOU  \\
\midrule
OneGAN                              & 20.5 & 30.7 & 55.5    \\
no loss $\mathcal{L}_{R_M}$         & 97.2 & 19.5 & 35.3    \\
no loss $\mathcal{L}_{\text{VAE}}$  & 44.1 & 18.5 & 39.6    \\
no loss $\mathcal{L}_{\text{PER}}$  & 25.5 & 24.1 & 53.0   \\
\bottomrule
\end{tabular}\\
(a)&(b)&(c)\\
\end{tabular}
\end{adjustbox}
\vspace{-.70cm}
\end{table*}

We train the network for 600,000 iterations, with batch size 20. All sub-networks are optimized using Adam \cite{adam}, with lr=2e-4. Phase I duration is 200,000 iterations and Phase II 400,000. Within Phase II, we start with training only on fake images and real image reconstruction starts after another 200,000 iterations.

We evaluate our model on various tasks against the state of the art methods. Since no other model can solve all these tasks, we evaluate against different methods in each task. Depending on availability, some baselines were pre-trained models released by the authors and some were trained from scratch with the authors' official code and instructions.

\noindent{\bf Datasets\quad}\label{sec:datasets}
We evaluate our model with three datasets of fine-grained categorization.
\textbf{Caltech-UCSD Birds-200-2011 (Birds)}~\cite{WahCUB_200_2011}: This dataset consists of 11,788 images of 200 classes of birds, annotated with bounding boxes and segmentation masks.
\textbf{Stanford Dogs (Dogs)}~\cite{KrauseStarkDengFei-Fei_3DRR2013}: This dataset consists of 20,580 images of 120 classes of dogs, annotated with bounding boxes. For evaluation, target segmentation masks were generated by a pre-trained DeepLabV3~\cite{chen2017rethinking} model on the COCO~\cite{lin2014microsoft} dataset. The pre-trained model was acquired from the gluoncv toolkit~\cite{gluoncvnlp2019}.
\textbf{Stanford Cars (Cars)}~\cite{KhoslaYaoJayadevaprakashFeiFei_FGVC2011}: This dataset consists of 16,185 images of 196 classes of cars, annotated with bounding boxes. Segmentation masks were generated as above with the pre-trained DeepLabV3 model.

Similarly to FineGAN, before training the model, we produced a background subset by cutting background patches with the bounding boxes. In addition to FineGAN, the bounding boxes were not used in any other way to train our method and we made sure that no image was used for both foreground and background examples. This was done by splitting the dataset in a 80/20 ratio, and use the larger subset as foreground $X_c$ and only the smaller subset for background $X_{bg}$.

Due to the different size of classes in each dataset, there is also a different size of child and parent classes in the design for each dataset. Birds: $N_C=200, N_P=20$, Dogs: $N_C=120, N_P=12$, Cars: $N_C=196, N_P=14$.

\noindent{\bf Image generation\quad}
We compare our image generation results to FineGAN~\cite{singh2018finegan} and StackGANv2~\cite{zhang2018stackgan++}, by relying on an InceptionV3 fine-tuned on each dataset.
We evaluate our method in both IS~\cite{gantricks} and FID~\cite{fid}. In addition, we measure the conditional variants of these metrics (CIS, CFID), as presented in \cite{benny2020evaluation}. The conditional metrics measure the similarity between real and fake images within each class, which cannot be measured by the unconditional metrics.

Our results, reported in Tab.~\ref{table:generation} show that OneGAN outperforms in both conditional and unconditional image generation metrics. In unconditional generation, our method and the baselines performed roughly the same, since the generators are very similar. In conditional generation, our method improves on the baseline by a large margin. StackGANv2 was the worst performing model, followed by FineGAN. This suggest that the mask-based generation, that FineGAN and our method rely on, generates a stronger conditioning on the object in the image. In addition, our multi-path training method improves conditional generation further, as is shown in the ablation tests. For illustration of conditional generation, see Fig.~\ref{fig:conditional}.


\noindent{\bf Unsupervised foreground segmentation\quad}
We compare our mask prediction from the reconstruction path to the real foreground mask. We evaluate according to IOU and DICE scores.
We compare against three baselines, ReDO~\cite{chen2019unsupervised}, WNet~\cite{xia2017w} and UISB~\cite{kanezaki2018unsupervised} which are trained for each dataset separately, and a third one, IIC-seg~\cite{ji2019invariant}, which was trained on coco-stuff and coco-stuff-3 (a subset). While coco-stuff is a different dataset than the ones we used, it contains all the relevant classes. ReDO and WNet produce a foreground mask which we compare to the ground-truth similarly to how we evaluate our model. UISB is an iterative method that produces a final segmentation with a varying number of classes between 2 and 100. We iterated UISB on each image 50 times. The output was usually between 4-20 classes. Since there is no labeling of the foreground or background classes, this method cannot be immediately used for this task. In order to get an evaluation, we look for each image for the class that has the highest IOU with the ground-truth foreground. The rest of the classes are merged to a single background class. We then repeat with a single background class and the rest merged into foreground. Finally, taking the best out of the two options, each obtained by using an oracle to select out of many options, which provides a liberal upper bound on the performance of UISB. IIC also produces a multi-class segmentation map, we use it in the same way we use UISB by taking the best class for either background or foreground in respect to IOU. IIC has 2-headed output, one for the main task and one for over-clustering. For coco-stuff trained IIC, we look for the best mask in one of the 15 classes of the main head. For coco-stuff-3 trained IIC, the main head is trained to cluster sky/ground/plants, so we look for the best mask in one of the 15 classes of the over-clustering head. FineGAN cannot perform segmentation, since it does not have a reconstruction path. But we added an additional baseline by training FineGAN with our encoders to allow such path. The results in Table.~\ref{table:segmentation} show that our method outperforms all the baselines. The ablation show that the biggest contribution comes from the reconstruction path and the multi-phase scheduling.

\noindent{\bf Unsupervised clustering\quad}
We compare our method against JULE~\cite{yang2016joint}, DEPICT~\cite{ghasedi2017deep} and IIC-cluster~\cite{ji2019invariant}. In addition, we added the baselines of StackGANv2 and FineGAN trained with our encoders. In this task, we evaluate how well the encoders are capable of clustering real images. The results show that OneGAN outperforms the other methods for both Birds and Dogs datasets. For Cars, our model was second after DEPICT. By looking at the generated images, this can be explained by the fact our method clusters the cars based more on color and less on car model. This aligns with the conditional generation scores, where the scores for Cars were lower than for the other datasets. 

\noindent{\bf Image to image translation\quad}
To further evaluate our model, we show its capability to transfer an input image to a target category. The results can be seen in Fig.~\ref{fig:transfer}. Even though our model was never trained on this task, the disentanglement between the shape and the texture enables this translation simply by passing a different child code during reconstruction. By selecting different child codes, we can manipulate the appearance of the object to any of the child categories. In contrast, FineGAN and StackGANv2 are unable to perform this task correctly as there is no learned disentanglement in StackGANv2's case and no bypass connection in FineGAN's case to allow good reconstruction.

\noindent{\bf Object removal and inpainting\quad}
Through the reconstruction task, our model is also capable of performing automatic object removal and background reconstruction, see Fig.~\ref{fig:reconstruction}. In contrast to other known method for inpainting, due to the lack of perfect ground-truth mask, our model does not only fill the missing pixels but fully reconstructs the background image. As a result, the background image is not identical to the original background, but it is semantically similar to it. we compare our method with previous work in the supplementary.

\noindent{\bf Ablation study\quad} In Tab.~\ref{table:generation},\ref{table:segmentation}, we provide multiple versions of our method for ablation. In the version without real reconstruction, we only add fake image reconstruction in Phase II, meaning that real images did not pass through the network during training. Another variant employs only the first phase of training. Finally, a third variant trains without multi-phase scheduling. These tests show the contribution of the multiple paths and the multi-phase scheduling. In Tab.~\ref{table:ablation}, we provide an extensive ablation study on three aspects. In (a), we compared layer and instance normalization~\cite{ulyanov2016instance} methods in the generators. Our ``GLU layer normalization'' outperformed all other options. In (b), we turned of intersection modules between encoders and generators. The experiment shows that these models strongly improve the CIS, which explains why our method outperformed FineGAN and StackGANv2 in conditional generation. In (c), we evaluated the contribution of selected novel losses, which affected all scores. Together, all these experiments show the contribution of the proposed novelties in our method.

\section{Conclusions}

By building a single model to handle multiple unsupervised tasks at once, we convincingly demonstrate the power of co-training, by surpassing the performance of the best in class methods for each task. This capability is enabled by a complex architecture with many sub-networks. Considering biological visual system, one can expect future architectures to be complex and to contain multiple pathways between the various modules. However, supporting this complexity during training is challenging. We introduce a mixup module that integrates multiple pathways in a homogenized manner and a multi-phase training, which helps to avoid some tasks dominating over the others.

\clearpage
\section*{Acknowledgement}
This project has received funding from the European Research Council (ERC) under the European Unions Horizon
2020 research and innovation programme (grant ERC CoG
725974).
%
%
\bibliographystyle{splncs04}
\bibliography{egbib}

\clearpage
\appendix
\section{Summary of notation}

For convenience, in Tab.~\ref{tab:overview} we provide a complete listing of the notation used in our paper.

\section{Regularization}

Due to lack of space in the main text, we include the regularization loss terms as part of the supplementary. 

During generation, we apply regularization on the latent vectors and on the mask image. The former serves to bound the range of the values to be close to the axis center and to be closely grouped.
\begin{equation*}
\mathcal{L}_{R_v} = || v_p ||^2_2 + || v_c ||^2_2 + || v_{bg} ||^2_2
\end{equation*}

The regularization on the mask serves to direct the model to utilize the mask efficiently, with a balanced and decisive representation of background and foreground. 
For mask $I_m \in [0,1]^{H,W}$, with $H,W$ the height and width of the mask. The first regularization term balances the mask value around the value of half.
\begin{equation*}
\mathcal{L}_{M_B} = | \frac{1}{HW} (\sum_{i,j}^{HW} (I_m)_{i,j}) - \frac{1}{2} |
\end{equation*}

The second regularization term aims to make the masks more decisive. It is better described when the mask is between [-1,1], so we define $I'_m= 2\cdot I_m -1$. In the ideal case, all pixels are either 1 or -1 (either background of foreground), therefore, if we assume a balanced distribution, for each mask the average value of $\max(0,I'_m)$ is 0.5 and of $\min(0,I'_m)$ it's -0.5, since half of the pixels are zeroed in each term. This is the decisiveness regularization.
\begin{equation*}
\mathcal{L}_{M_D} = |\frac{1}{HW} (\sum_{i,j}^{HW}\max(0,(I_m')_{i,j})) - \frac{1}{2}| +
                    |\frac{1}{HW} (\sum_{i,j}^{HW}\min(0,(I_m')_{i,j})) + \frac{1}{2}|)
\end{equation*}

Together, the mask regularization loss is:
\begin{equation*}
\mathcal{L}_{R_M} = \mathcal{L}_{M_B} + 0.1 \cdot \mathcal{L}_{M_D}
\end{equation*}

\section{Sub-networks architecture}
In this section, we describe the details of each sub-network described in 
the main paper. The layers of each sub-network are listed in the tables Tab.~\ref{table:arch_Gbg}--\ref{table:arch_Dc}, with some modules that are frequently used listed in Tab.~\ref{table:arch_general}. The majority of the networks are sequential. When more complicated connections are present, the input and output notations are there to guide the flow.

\section{Additional illustrations}

\subsection{Conditional generation}

We supply more conditionally generated images to further demonstrate the conditional generation performance. We use generation conditioned on both very different classes to highlight the broad coverage of the representation and very similar classes to show the high sensitivity to detail.

\begin{figure*}[t]
\centering
\includegraphics[width = \textwidth,trim={0 650 0 0}, clip]{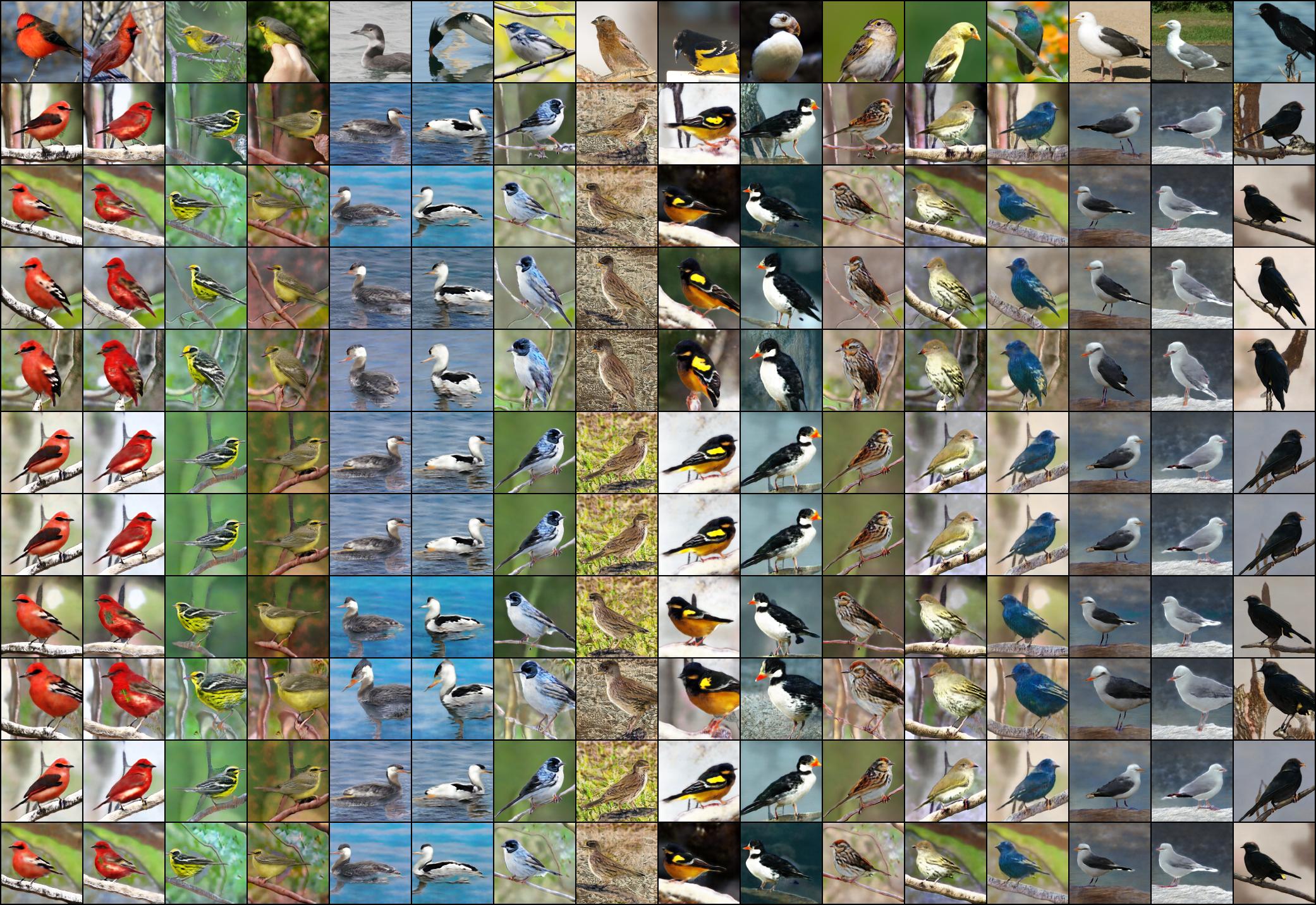}
\includegraphics[width = \textwidth,trim={0 650 0 0}, clip]{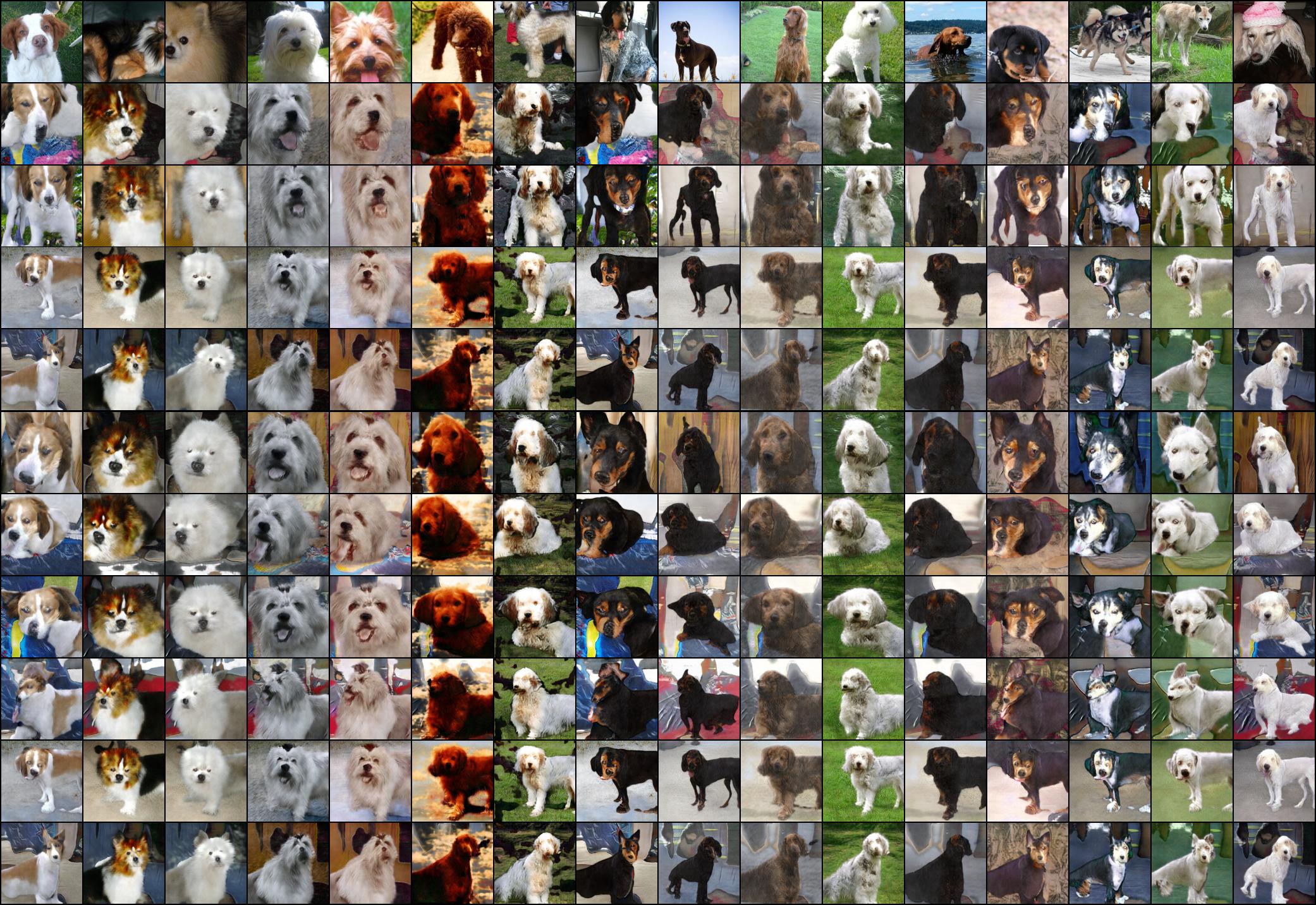}
\includegraphics[width = \textwidth,trim={0 650 0 0}, clip]{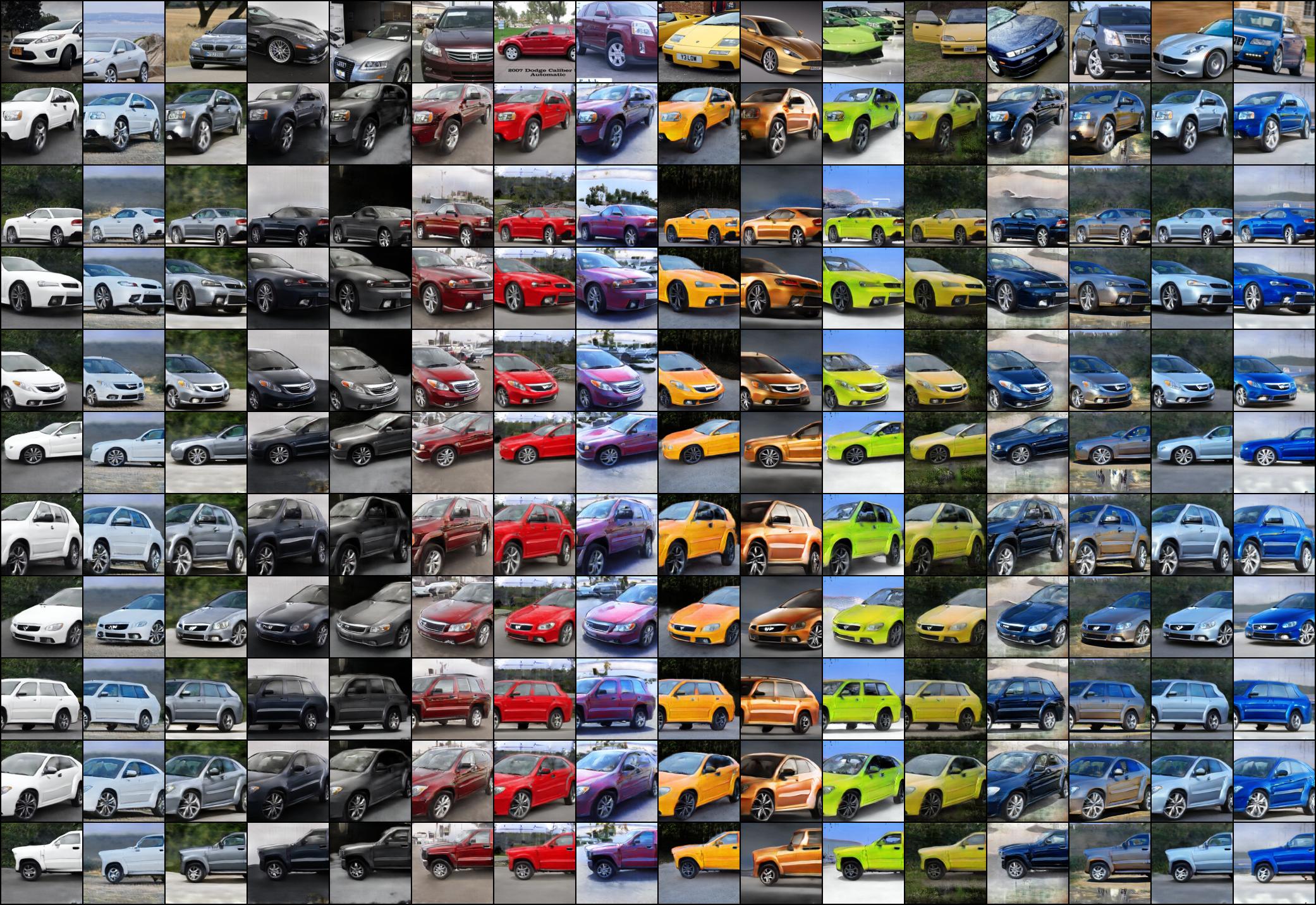} \\
\caption{Conditional Image Generation. From top to bottom: (i) real image, (ii-vi) generation of images with the encoded parent and child codes and a different vector $z$ per row.}
\label{fig:conditional_ex}
\end{figure*}

In Fig.~\ref{fig:conditional_ex}, the images are obtained by generating five different images per reference image in the top row. To achieve these results, our model has to perform two tasks. First, it has to be able to detect the child and parent classes under which the object is represented. Second, it needs to be able to generate a similar looking object with the predicted classes. The success in this task is evidence for both the generation and clustering capabilities of the model. 

In the figure, each column shows a real image, followed by five generated images conditioned on the first image in respect to category. Additionally, in each row, all images are generated with the same $z$, showing how non-categorical information is consistent across the different categories and how the pose is disentangled from the shape category.

For both birds and dogs, we can see that the generated images have very similar properties to their conditioned image. In both cases, we can see the large coverage of different classes and the fine detailed differences between similar classes. For cars, we can see that that the generated images apply the same color, but the car shape is changing, indicating that the model has categorized the cars by their color and not by their model.

\subsection{Reconstruction}

We supply more images to show the reconstruction path with the resulting reconstructed images, reconstructed backgrounds, and segmentation masks. 

\begin{figure*}[t]
\centering
\includegraphics[width = \textwidth,trim={0 0 0 0}, clip]{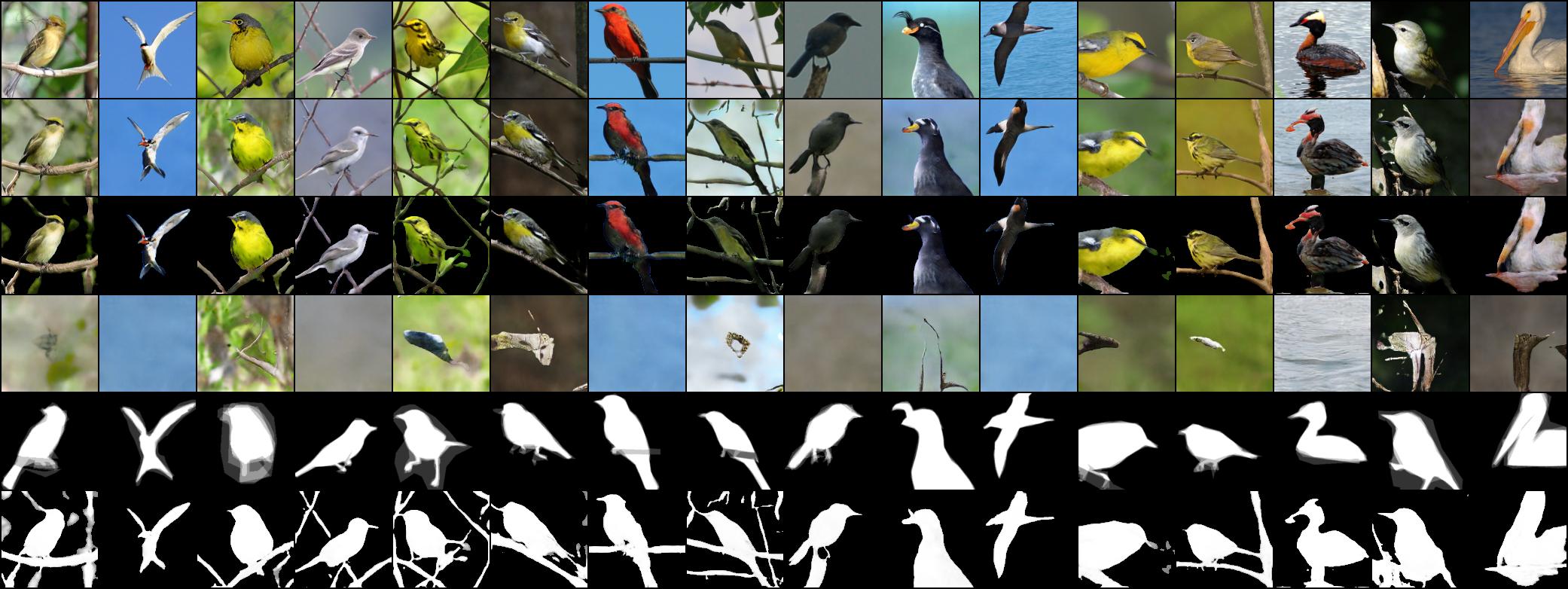}
\includegraphics[width = \textwidth,trim={0 0 0 0}, clip]{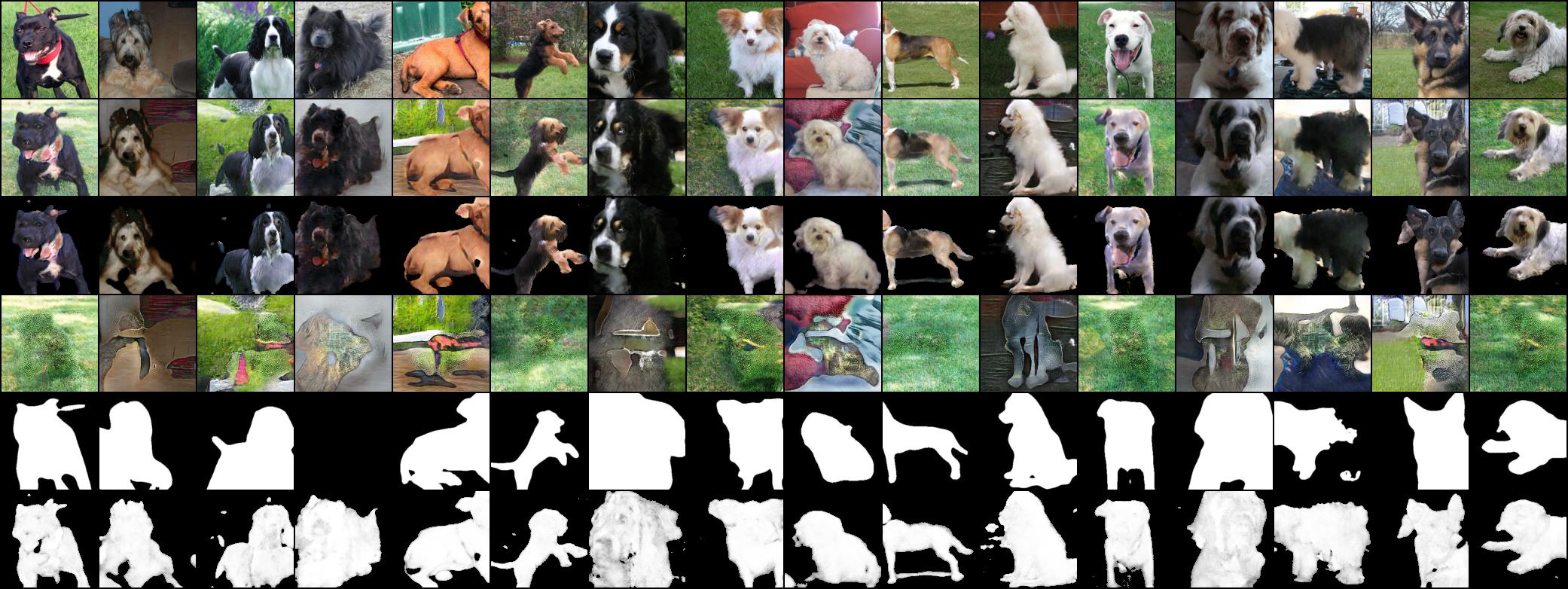}
\includegraphics[width = \textwidth,trim={0 0 0 0}, clip]{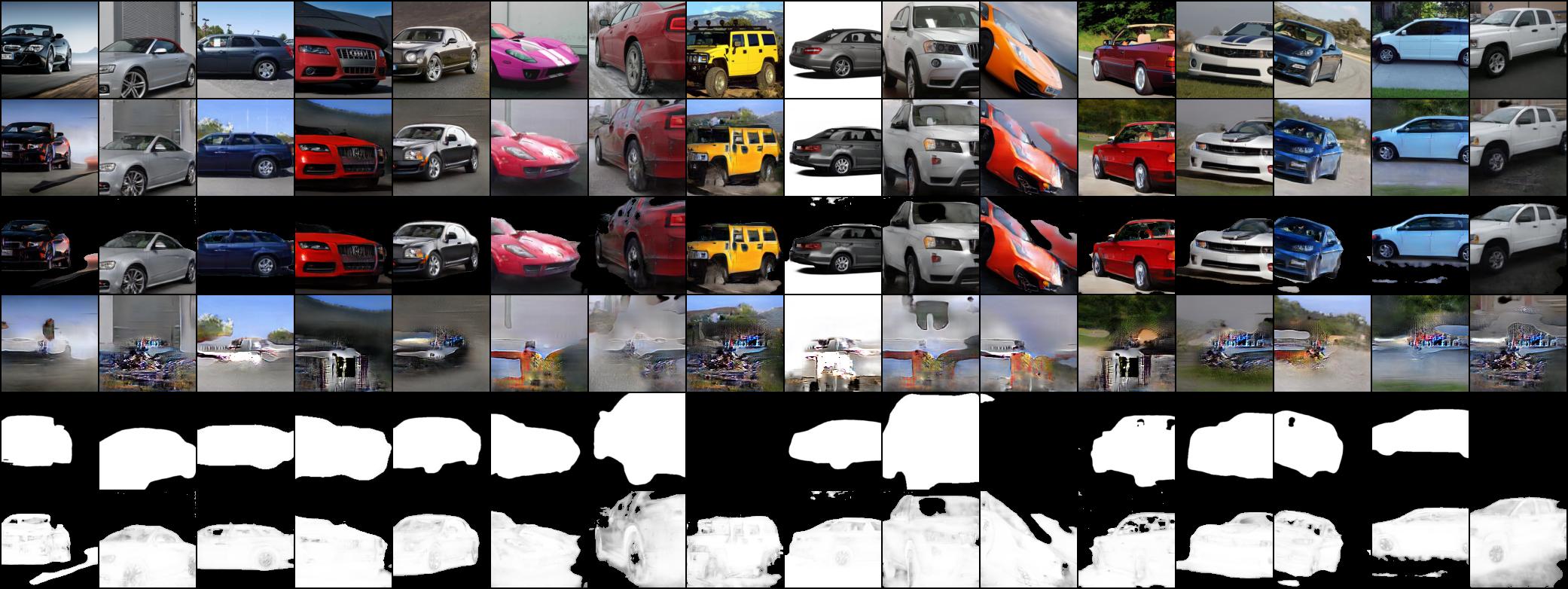} \\
\caption{Image Reconstruction. From top to bottom: (i) real image, (ii) reconstructed image, (iii) reconstructed foreground, (iv) reconstructed background, (v) ground-truth foreground mask, (vi) predicted foreground mask.}
\label{fig:reconstruction_ex}
\end{figure*}

In Fig.~\ref{fig:reconstruction_ex}, the results show the images generated by the reconstruction process. The generated mask shows the model's ability to detect and segment the object, the background image shows the model's ability to repaint the background, and the foreground image shows the model's ability to detect and reconstruct conditioned on the object class. 

The segmentation works under many different poses, sizes, and backgrounds. For dogs and cars, it can be seen that our model is sometimes better than the ``ground-truth'' masks, which were generated by a pre-trained network. The background repainting works well in the majority of cases, but we do notice that some backgrounds work better than others. The challenges are mostly noticeable in the cars dataset, where there was the smallest amount of background patches available in $X_{bg}$, leading to a less powerful background discriminator. Subsequently, the performance of the background generation was affected.

\subsection{Image to image translation}

\begin{figure*}[t]
\centering
\includegraphics[width = \textwidth,trim={0 520 0 0}, clip]{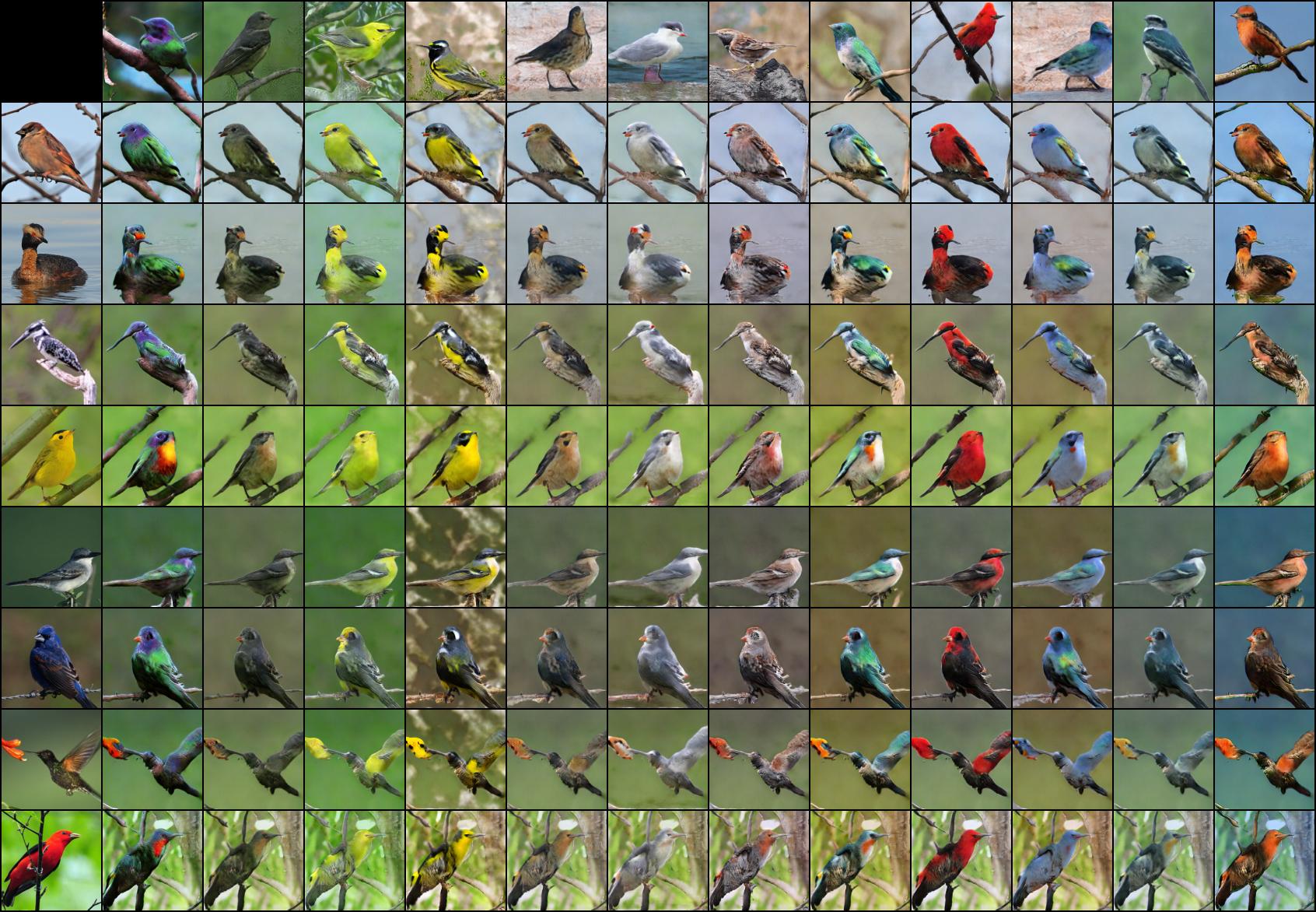}
\includegraphics[width = \textwidth,trim={0 780 0 0}, clip]{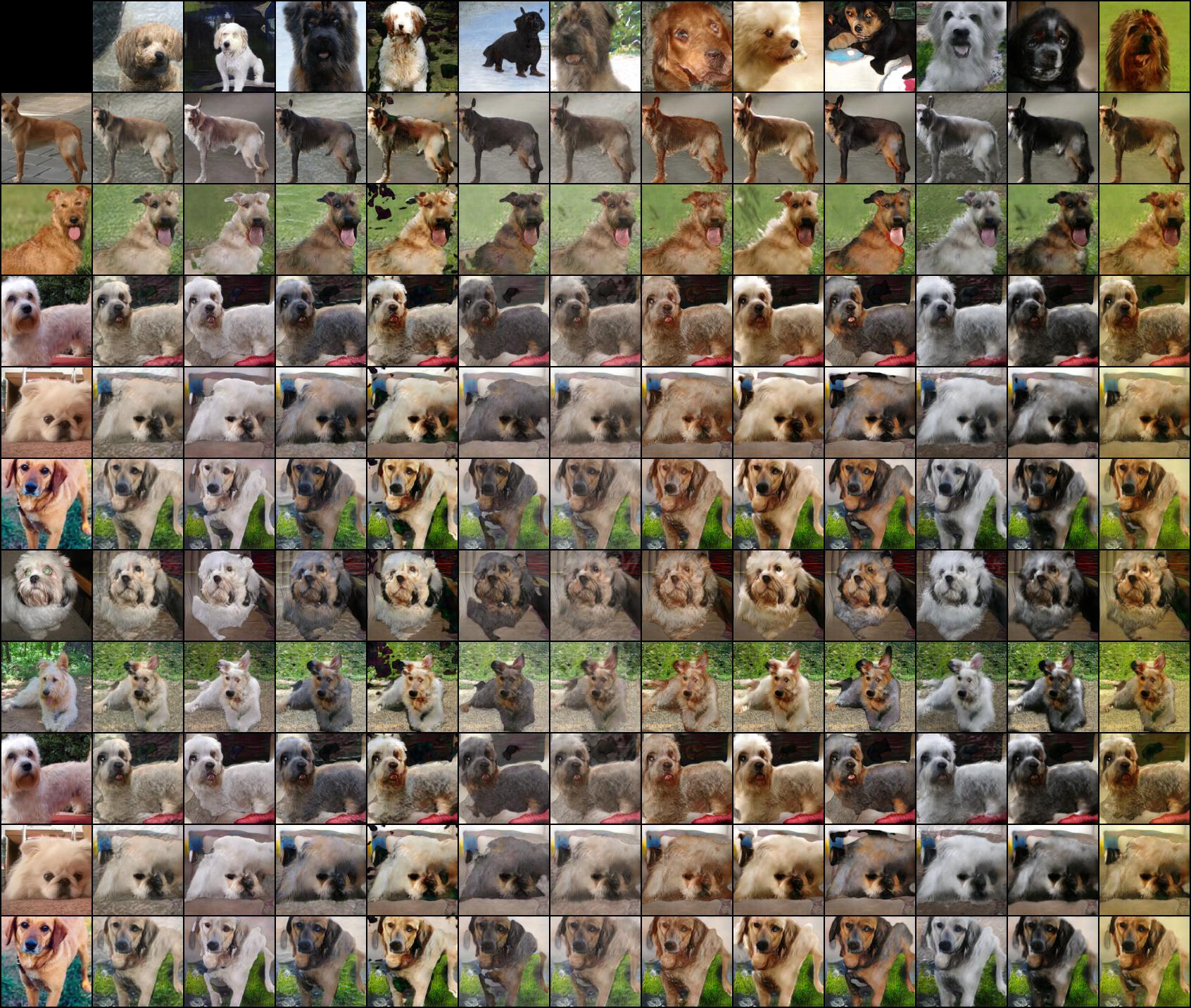}
\includegraphics[width = \textwidth,trim={0 780 130 0}, clip]{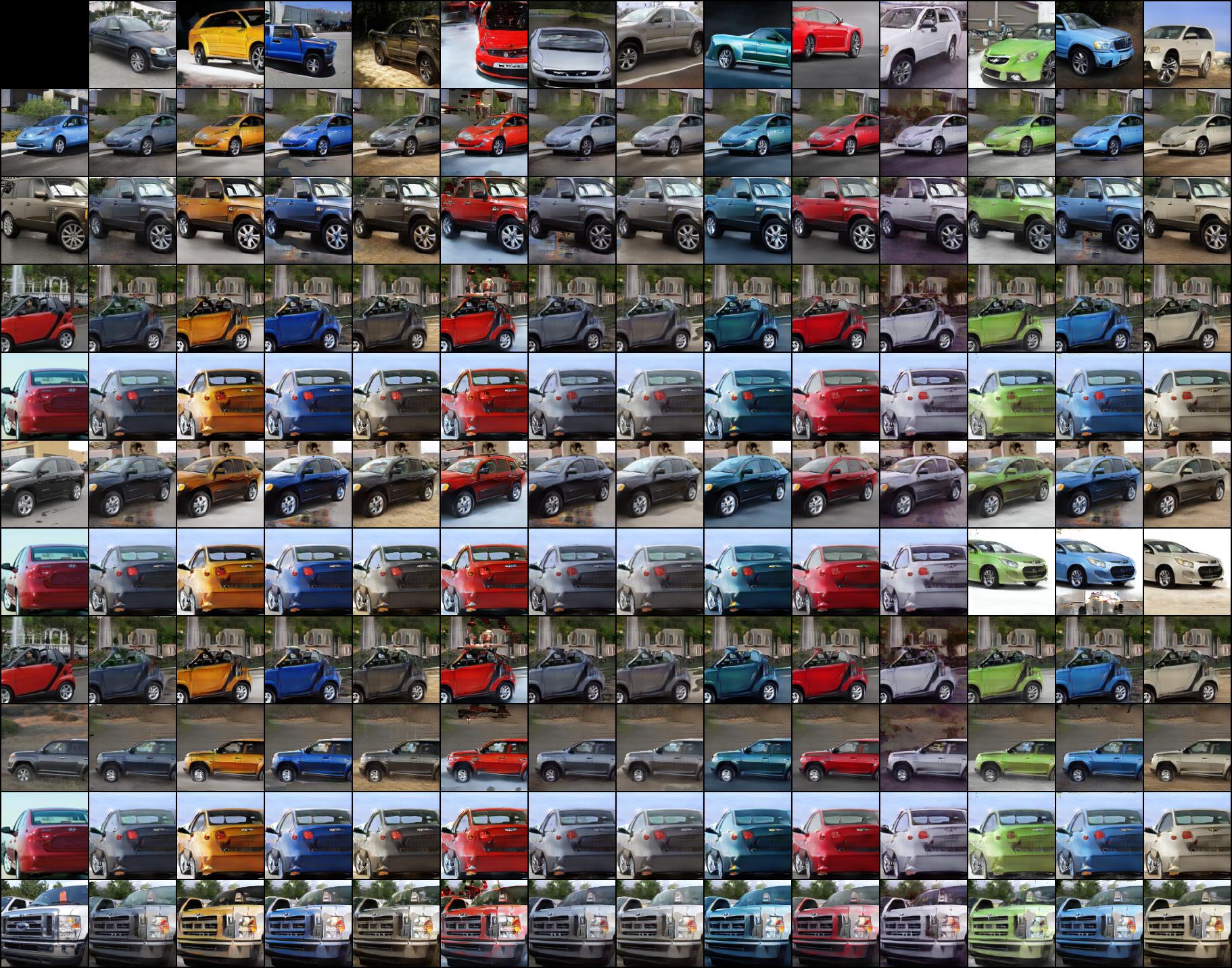} \\
\caption{Image to Image Translation. From left to right: (i) real image, (ii-xiii) reconstructed images when the child code $e_c$ in each column is switched with a code from a selected category represented by the top image.}
\label{fig:translation_ex}
\end{figure*}

We supply more images, Fig.~\ref{fig:translation_ex}, to show the image to image translation capability of the model. We show that the disentanglement that emerged from the design allows manipulation of the reconstructed image by replacing the child code with a code from an arbitrary category. We can see that not only do the objects in the images change appearance, but the change is consistent across different images, while the background is mostly unaffected. 

In birds, we can see that the generator is usually able to detect the different parts of the birds (wings, head, beak) and apply the correct color manipulation to the correct area. Furthermore, the color manipulation works on birds of different shapes and different original colors. The background is sometimes slightly altered, first, because it is regenerated every time, and second, because the foreground mask is soft and sometimes applies a slight manipulation on the background as well. In dogs, the manipulation is less effective, but it is noticeable and is correlated to the applied category. In cars, we can see the color manipulation for many different colors. We can also notice how the background is mostly unchanged and that the manipulation is applied correctly on the car chassis and not on other parts like windows, tires and lights.

\subsection{Background inpainting}

Due to space limits, we could not address all tasks in the main paper. As an intermediate step of the reconstruction task, our model also performs a side-task of object foreground extraction and background inpainting. The model first detects the foreground in the image and produces a segmentation mask. Then, with the mask, the background is encoded and reconstructed. Because the mask is a prediction and not a ground-truth, the model cannot only fill the masked pixels with background texture, but has to assume that the mask was not perfect and reconstruct the entire image. The drawback of this method is that the background is not always identical to the source in the background area, but the benefit is that the object is fully removed even when it is not fully covered by the mask.

We compare our model against images produced with Deep-Image-Prior (DIP; Ulyanov et al., CVPR 2018). There are two variants. In the fist, DIP receives the ground-truth mask and in the second, the predicted mask is given. DIP optimizes its network for 1000 steps on the input image. 

\begin{figure*}[t]
\centering
\includegraphics[width = \textwidth,trim={0 780 0 0}, clip]{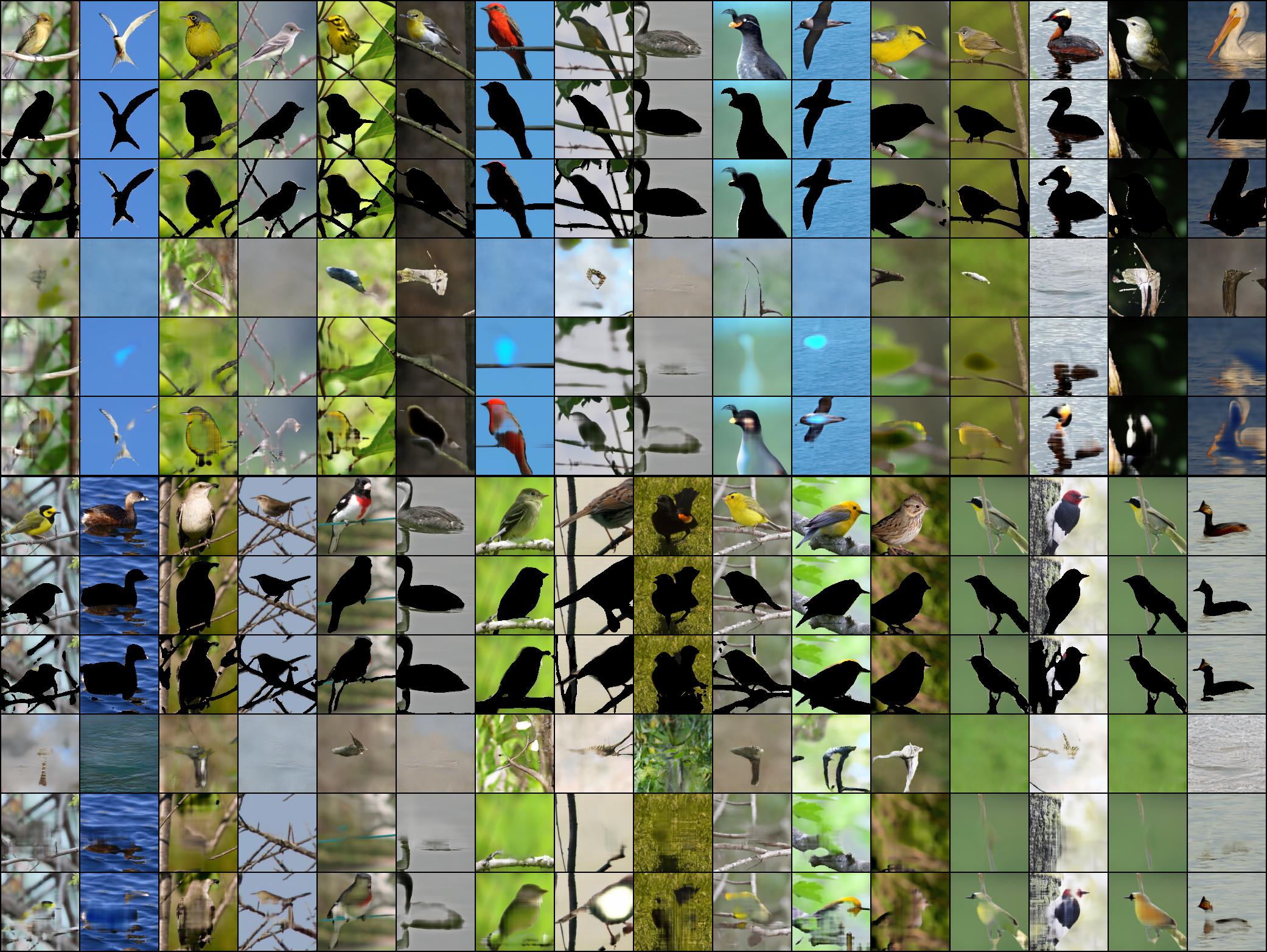}
\includegraphics[width = \textwidth,trim={0 780 0 0}, clip]{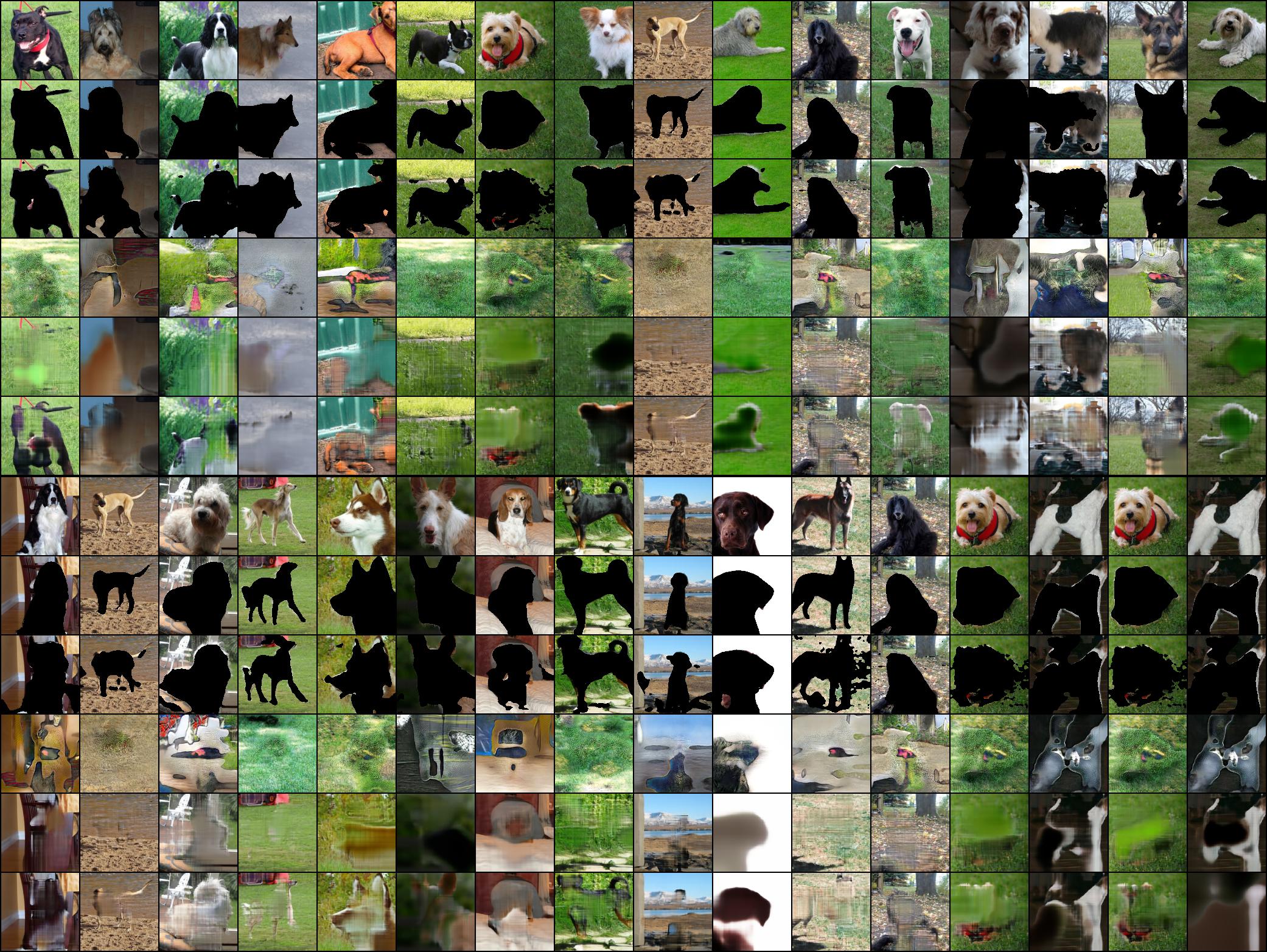}
\includegraphics[width = \textwidth,trim={0 780 0 0}, clip]{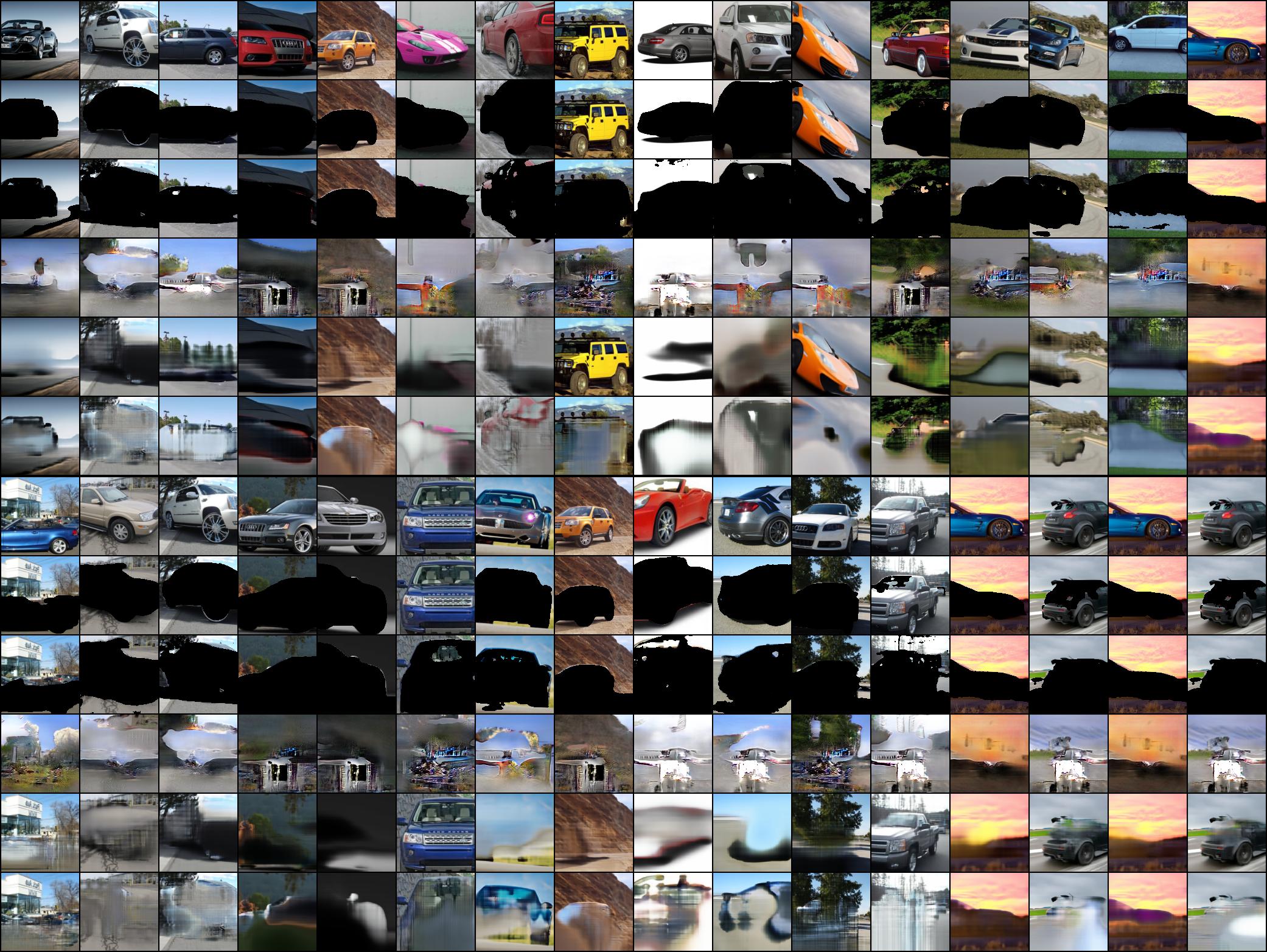} \\
\caption{Background Inpainting. From top to bottom: (i) original image, (ii) image masked with real mask, (iii) image masked with predicted mask, (iv) OneGAN, (v) DIP with real mask, (vi) DIP with predicted mask.}
\label{fig:inpainting_ex}
\end{figure*}

The results can be seen in Fig.~\ref{fig:inpainting_ex}. It can be observed that DIP works relatively well when using a perfectly covering mask, but fails when the mask is not perfect and does not fully cover the object. In contrast, our model suffers less when the mask is not perfect. We can also see that our model does not exactly inpaints the background but actually repaints it, which usually results in a slightly different background even where the image was not masked, but as we mentioned above, it may be beneficial when the mask is not perfect. Finally, our model performs the inpainting task in a single forward path instead of 1000 iterations of DIP.

\begin{table*}[t]
\centering
\caption{The components of the OneGAN model}\label{tab:overview}
  \begin{tabular}{llll}
    \toprule
    & Symbol     & Description     & Computed as (or a comments) \\
    \midrule
    \parbox[t]{3mm}{\multirow{3}{*}{\rotatebox[origin=c]{90}{Variables~~~}}}
    & $\phi_c \in [1,N_C]$ & child class & \\
    & $\phi_p \in [1,N_P]$ & parent class & \\
    & $e_c \in \{0,1\}^{N_C}$ & child class one-hot vector (style) & $e_c[i] = \delta_{i,\phi_c}$ \\
    & $e_p \in \{0,1\}^{N_P}$ & parent class one-hot vector (shape) & $e_p[i] = \delta_{i,\phi_p}$ \\
    & $e_{bg} \in \{0,1\}^{N_P}$ & background one-hot vector & $e_{bg} = e_p$ \\
    & $z \in \mathbb{R}^{d_z}$ & pose code & $z[i] \sim \mathcal{N}(0,1)$ \\
    & $v_c \in \mathbb{R}^{d_c}$ & style code vector & $v_c = V_{c_0}(e_c)$ \\
    & $v_p \in \mathbb{R}^{d_p}$ & shape code vector & $v_p = V_{p_0}(e_p)$ \\
    & $v_{bg} \in \mathbb{R}^{d_{bg}}$ & background code vector & $v_{bg} = V_{bg_0}(e_{bg})$ \\
    & $A_{fg}$ & foreground pre-image & $A_{fg} = G_{{fg}_0}(v_p, z)$ \\
    & $A_{bg}$ & background pre-image & $A_{bg} = G_{{bg}_0}(v_{bg}, z)$ \\
    & $I_{m}$ & foreground mask & \\
    & $I_{fg}$ & foreground image & $(I_{fg}, I_m) = G_{{fg}_2}( G_{{fg}_1}(A_{fg}, v_p), v_c)$ \\
    & $I_{bg}$ & background image & $I_{bg} = G_{{bg}_1}(A_{bg})$ \\
    & $I$ & full image & $I = I_{bg} \circ (1-I_m) + I_{fg} \circ I_m$\\
    & $B_{fg}$ & foreground bypass & $B_{fg} = E_{{p}_1}(I)$ \\
    & $B_{bg}$ & background bypass & $B_{bg} = E_{{bg}_1}(I, I_m)$ \\
    & $X_{c}$ & image domain & \\
    & $X_{bg}$ & background image domain & \\
    \midrule
      
    \parbox[t]{3mm}{\multirow{3}{*}{\rotatebox[origin=c]{90}{Networks~~~}}} 
    & $V_c$ & embedding LUT of child class & \\
    & $V_p$ & embedding LUT of parent class & \\
    & $V_{bg}$ & embedding LUT of background & \\
    & $G_{fg}$ & foreground generator  &  $G_{{fg}_2}( G_{{fg}_1}(G_{{fg}_0}(v_p , z), v_p) , v_c)$ \\
    & $G_{bg}$ & background generator & $G_{{bg}_1}( G_{{bg}_0}(v_{bg} , z))$ \\
    & $E_c$ & style encoder & $E_c(I)$ \\
    & $E_p$ & content encoder & $E_p(I)$ \\
    & $E_{bg}$ & content encoder & $E_{bg}(I, I_m)$ \\
    & $D_c$ & image discriminator & \\
    & $D_{bg}$ & background discriminator & \\
    \midrule
    
    \parbox[t]{3mm}{\multirow{3}{*}{\rotatebox[origin=c]{90}{Parameters~~~}}}
    & $N_C$ & number of child classes & Depends on the dataset. \\
    & $N_P$ & number of parent classes & $N_P < N_C$, Depends on the dataset. \\
    & $d_z$ & dimensionality of pose code & $d_z = 100$ \\
    & $d_c$ & dimensionality of style code & $d_c = 32$ \\
    & $d_p$ & dimensionality of shape code & $d_p = 16$ \\
    & $d_{bg}$ & dimensionality of background code & $d_{bg} = 32$ \\
    & $H,W$ & size of image & $H = W = 128$ \\
  
    \bottomrule
  \end{tabular}
\end{table*}

\clearpage

\clearpage

\begin{table}[ht]
\caption{General modules}\label{table:arch_general}
\centering
\begin{tabular}{lccc}
\toprule
Module                      & layers & input & output\\ 
\midrule
\multirow{4}{*}{GLU-LNorm}  & ChannelSplit  & $x$           & $x_L, x_R$   \\
                            & LayerNorm     & $x_L$         & $x_L'$   \\
                            & Sigmoid       & $x_R$         & $x_R'$ \\
                            & Multiply      & $x_L',x_R'$   & - \\
\midrule
\multirow{3}{*}{UPBlk}      & Upsample2d($S$/2, $S$)        & -     & - \\
\multirow{3}{*}{($c_i,c_o,S$)} 
                            & K3P1Conv2d($c_{i},2c_{o}$)    & -     & - \\
                            & GLU-LNorm                     & -     & - \\ 
\midrule
\multirow{3}{*}{DOWNBlk}    & K4S2P1Conv2d($c_i$,$c_o$)     & -     & - \\
\multirow{3}{*}{($c_i,c_o$)}
                            & LayerNorm                     & -     & - \\ 
                            & lReLU(0.2)                    & -     & - \\       
\midrule                            
\multirow{5}{*}{RESBlk0}    & K3P1Conv2d($c_i$,$2c_i$)      & $x$   & - \\ 
\multirow{5}{*}{($c_i$)}
                            & GLU-LNorm                     & -     & -     \\
                            & K3P1Conv2d($c_i$,$2c_i$)      & -     & -     \\ 
                            & GLU-LNorm                     & -     & $d$   \\
                            & Add                           & $x,d$ & -     \\
\midrule                            
\multirow{6}{*}{RESBlk}     & K3P1Conv2d($c_i+d$,$2c_i$)  & -  & -    \\
\multirow{6}{*}{($c_i,d,c_o$)}
                            & GLU-LNorm                 & -  & -    \\ 
                            & RESBlk0($c_i$)            & - & -     \\
                            & RESBlk0($c_i$)            & - & -     \\
                            & K3P1Conv2d($c_i$,2$c_o$)  & -  & -    \\
                            & GLU-LNorm                 & -  & -    \\ 
\bottomrule
\end{tabular}
\end{table}

\begin{table}[ht]
\caption{Background Generator $G_{bg}$}\label{table:arch_Gbg}
\centering
\begin{tabular}{lccc}
\toprule
Module                          & layers & input & output\\ 
\midrule
\multirow{1}{*}{$V_{bg}$}       & Linear($N_P$, $d_{bg}$)  & $e_{bg}$  & $v_{bg}$   \\
\cmidrule(lr){2-4}
\multirow{5}{*}{$G_{bg_0}$}     & Linear($d_{bg}+d_z$, 32768)  & $v_{bg},z$  & -   \\
                                & Reshape(2048,4,4) & -  & - \\
                                & GLU-LNorm         & -  & -      \\ 
                                & UPBlk(1024,512,8) & -  & -      \\ 
                                & UPBlk(512,256,16) & -  & $A_{bg}$      \\
\cmidrule(lr){2-4}
\multirow{4}{*}{$G_{bg_1}$}     & UPBlk(256,128,32) & $A_{bg}$  & -   \\
                                & UPBlk(128,64,64)  & -  & -     \\ 
                                & UPBlk(64,32,128)  & -  & -      \\ 
                                & K3P1Conv2d(32,3) + tanh & -  & $I_{bg}$      \\                                 
\bottomrule
\end{tabular}
\end{table}

\begin{table}[ht]
\caption{Foreground Generator $G_{fg}$}\label{table:arch_Gfg}
\centering
\begin{tabular}{lccc}
\toprule
Module                          & layers & input & output\\ 
\midrule
\multirow{1}{*}{$V_{p}$}        & Linear($N_P$, $d_{p}$)  & $e_{p}$  & $v_{p}$   \\
\midrule
\multirow{1}{*}{$V_{c}$}        & Linear($N_C$, $d_{c}$)  & $e_{c}$  & $v_{c}$   \\
\midrule
\multirow{5}{*}{$G_{fg_0}$}     & Linear($d_{p}+d_z$, 32768)  & $v_{p},z$  & -   \\
                                & Reshape(2048,4,4) & -  & - \\
                                & GLU-LNorm         & -  & -      \\ 
                                & UPBlk(1024,512,8) & -  & -      \\ 
                                & UPBlk(512,256,16) & -  & $A_{fg}$      \\
\midrule
\multirow{3}{*}{$G_{fg_1}$}     & UPBlk(256,128,32) & $A_{fg}$  & -   \\
                                & UPBlk(128,64,64)  & -  & -        \\ 
                                & UPBlk(64,3,128)   & -  & $C_{fg_0}$  \\
\midrule
\multirow{4}{*}{$G_{fg_2}$}     & RESBlk(64,$d_p$,32)  & $C_{fg_0},v_p$  & $C_{fg_1}$ \\
                                & RESBlk(32,$d_c$,16)  & $C_{fg_1},v_c$  & $C_{fg_2}$ \\ 
                                \cmidrule(lr){2-4}
                                & K3P1Conv2d(16,3) + tanh & $C_{fg_2}$  & $I_{fg}$      \\
                                \cmidrule(lr){2-4}
                                & K3P1Conv2d(16,1) + sigmoid & $C_{fg_2}$  & $I_{m}$    \\
\bottomrule
\end{tabular}
\end{table}

\begin{table}[ht]
\caption{Style Encoder $E_c$}\label{table:arch_Ec}
\centering
\begin{tabular}{lccc}
\toprule
Module                          & layers & input & output\\ 
\midrule
\multirow{5}{*}{$E_{c_1}$}      & K4S2P1Conv2d(3, 64)   & $I$  & -   \\
                                & LayerNorm             & - & - \\
                                & lReLU(0.2)            & - & - \\
                                & DOWNBlk(64, 128)      & -  & -   \\
                                & DOWNBlk(128, 256)     & - & $H_c$   \\
\midrule
\multirow{12}{*}{$E_{c_0}$}      
                                & DOWNBlk(256, 512)         & $H_c$  & -   \\
                                & DOWNBlk(512, 1024)        & -  & -   \\
                                & K3P1Conv2d(1024, 1024)    & - & -   \\
                                & LayerNorm                 & - & - \\
                                & lReLU(0.2)                & - & - \\
                                & Reshape(16384)            & - & - \\
                                & Linear(16384, 512)        & - & -   \\ 
                                & LayerNorm                 & - & - \\
                                & lReLU(0.2)                & - & $h_c$ \\ 
                                \cmidrule(lr){2-4}
                                & Linear((512, $N_C$)   & $h_c$ & $\hat{e}_c$   \\
                                & Linear(512, $d_c$)    & $h_c$ & $\mu_c$   \\
                                & Linear(512, $d_c$)    & $h_c$ & $\sigma_c$   \\
\bottomrule
\end{tabular}
\end{table}

\begin{table}[ht]
\caption{Shape Encoder $E_p$}\label{table:arch_Ep}
\centering
\begin{tabular}{lccc}
\toprule
Module                          & layers & input & output\\ 
\midrule
\multirow{8}{*}{$E_{p_1}$}      & K4S2P1Conv2d(3, 64)   & $I$  & -   \\
                                & LayerNorm             & - & - \\
                                & lReLU(0.2)            & - & - \\
                                & DOWNBlk(64, 128)      & -  & -   \\
                                & DOWNBlk(128, 256)     & - & $H_p$   \\
                                \cmidrule(lr){2-4}
                                & K3P1Conv2d((256, 512) & $H_p$ & -   \\
                                & GLU-LNorm             & - & -  \\
                                & UPBlk(256,256)        & - & $B_{fg}$ \\
\midrule
\multirow{17}{*}{$E_{p_0}$}     & DOWNBlk(256, 512)         & $H_p$ & -   \\
                                & DOWNBlk(512, 1024)        & -  & -   \\
                                & K3P1Conv2d(1024, 1024)    & - & -   \\
                                & LayerNorm                 & - & - \\
                                & lReLU(0.2)                & - & - \\
                                & Reshape(16384)            & - & $h$ \\
                                \cmidrule(lr){2-4}
                                & Linear(16384, 512)        & $h$ & -   \\
                                & LayerNorm                 & - & - \\
                                & lReLU(0.2)                & - & $h_p$ \\
                                & Linear((512, $N_C$)       & $h_p$ & $\hat{e}_p$   \\
                                & Linear(512, $d_c$)        & $h_p$ & $\mu_p$   \\
                                & Linear(512, $d_c$)        & $h_p$ & $\sigma_p$   \\
                                \cmidrule(lr){2-4}
                                & Linear(16384, 512)        & $h$ & -   \\
                                & LayerNorm                 & - & - \\
                                & lReLU(0.2)                & - & $h_z$ \\ 
                                & Linear(512, $d_z$)        & $h_z$ & $\mu_z$   \\
                                & Linear(512, $d_z$)        & $h_z$ & $\sigma_z$   \\
\bottomrule
\end{tabular}
\end{table}

\begin{table}[ht]
\caption{Background Encoder $E_{bg}$}\label{table:arch_Ebg}
\centering
\begin{tabular}{lccc}
\toprule
Module                          & layers & input & output\\ 
\midrule
\multirow{6}{*}{$E_{bg_1}$}     & K4S2P1Conv2d(4, 64)   & $I, I_m$  & -   \\
                                & DOWNBlk(64, 128)      & -  & -   \\
                                & DOWNBlk(128, 256)     & - & $H_{bg}$   \\
                                & K3P1Conv2d((256, 512) & $H_{bg}$ & -   \\
                                & GLU-LNorm             & - & -  \\
                                & UPBlk(256,256)        & - & $B_{bg}$ \\
\bottomrule
\end{tabular}
\end{table}

\begin{table}[h]
\caption{Background Discriminator $D_{bg}$}\label{table:arch_Dbg}
\centering
\begin{tabular}{lccc}
\toprule
Module                          & layers & input & output\\ 
\midrule
\multirow{8}{*}{$D_{bg}$}       & DownSample2d(128, 126)  & $I$  & -   \\
                                & K4S2P0Conv2d(3, 64)  & -  & -   \\
                                & lReLU(0.2)  & -  & -      \\
                                & K4S2P0Conv2d(64, 128)  & -  & -   \\
                                & lReLU(0.2)  & -  & -      \\ 
                                & K4S4P0Conv2d(128, 256)  & -  & -   \\
                                & lReLU(0.2)  & -  & $H = D_{bg_C}(I)$      \\ 
                                \cmidrule(lr){2-4}
                                & K4S1P0Conv2d(256,1) & $H$ & $D_{bg_A}(I)$ \\
                                \cmidrule(lr){2-4}
                                & K4S1P0Conv2d(256,1) & $H$ & $D_{bg_B}(I)$ \\
\bottomrule
\end{tabular}
\end{table}

\begin{table}[ht]
\caption{Object Discriminator $D_c$}\label{table:arch_Dc}
\centering
\begin{tabular}{lccc}
\toprule
Module                          & layers & input & output\\ 
\midrule
\multirow{19}{*}{$D_{c}$}       & K4S2P1Conv2d(3, 64)   & $I$   & - \\
                                & LayerNorm             & -     & - \\
                                & lReLU(0.2)            & -     & - \\
                                & DOWNBlk(64, 128)      & -     & - \\
                                & DOWNBlk(128, 256)     & -     & - \\
                                & DOWNBlk(256, 512)     & -     & $D_{c_C}(I)$  \\
                                & DOWNBlk(512, 1024)    & -     & - \\  
                                & K3P1Conv2d(1024, 1024)& -     & - \\
                                & LayerNorm             & -     & - \\
                                & lReLU(0.2)            & -     & - \\
                                & Reshape(16384)        & -     & $H$ \\
                                \cmidrule(lr){2-4}
                                & Linear(16384, 512)    & $H$   & - \\
                                & LayerNorm             & -     & - \\
                                & lReLU(0.2)            & -     & - \\
                                & Linear(512, 1)        & -     & $D_{c_A}(I)$  \\
                                 \cmidrule(lr){2-4}
                                & Linear(16384, 512)    & $H$   & - \\
                                & LayerNorm             & -     & - \\
                                & lReLU(0.2)            & -     & - \\
                                & Linear(512, $N_C$)    & -     & $D_{c_B}(I)$  \\
\bottomrule
\end{tabular}
\end{table}
\end{document}